\documentclass{article}

\usepackage{arxiv}

\usepackage[utf8]{inputenc} 
\usepackage[T1]{fontenc}    
\usepackage[hidelinks]{hyperref}       
\usepackage{url}            
\usepackage{booktabs}       
\usepackage{amsfonts}       
\usepackage{nicefrac}       
\usepackage{microtype}      
\usepackage{amsmath}
\usepackage{cleveref}       
\usepackage{graphicx}
\usepackage[sort&compress,square,comma,numbers]{natbib}
\usepackage{doi}

\usepackage{placeins}
\usepackage{tabularx}
\usepackage[]{subfig}
\usepackage{booktabs}
\usepackage{multirow}
\usepackage{url}
\usepackage[export]{adjustbox}

\title{Learned Compression of Point Cloud Geometry and Attributes in a Single Model through Multimodal Rate-Control}

\date{}

\newif\ifuniqueAffiliation

\ifuniqueAffiliation 
\author{ \href{https://orcid.org/0000-0001-5566-7783}{\includegraphics[scale=0.06]{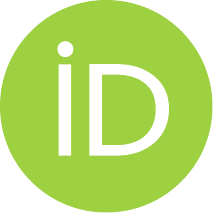}\hspace{1mm}Michael Rudolph}\\
	\texttt{rudolph@ikt.uni-hannover.de} \\
	\And
	\href{https://orcid.org/009-0001-1658-0762}{\includegraphics[scale=0.06]{orcid.pdf}\hspace{1mm}Aron Riemenschneider} \\
	\texttt{aron.riemenschneider@stud.uni-due.de} \\
	\And
	\href{https://orcid.org/0000-0000-0000-0000}{\includegraphics[scale=0.06]{orcid.pdf}\hspace{1mm}Amr Rizk} \\
	\texttt{amr.rizk@ikt.uni-hannover.de} \\
}
\else
\usepackage{authblk}

\newbox{\orcid}\sbox{\orcid}{\includegraphics[scale=0.06]{orcid.pdf}} 
\author[1,2]{%
	\href{https://orcid.org/0000-0001-5566-7783}{\usebox{\orcid}\hspace{1mm}Michael Rudolph \thanks{\texttt{michael.rudolph@ikt.uni-hannover.de}}}%
}
\author[1]{%
	\href{https://orcid.org/0009-0001-1658-0762}{\usebox{\orcid}\hspace{1mm}Aron Riemenschneider \thanks{\texttt{aron.riemenschneider@stud.uni-due.de}}}%
}
\author[2]{%
	\href{https://orcid.org/0000-0002-9385-7729}{\usebox{\orcid}\hspace{1mm}Amr Rizk \thanks{\texttt{amr.rizk@ikt.uni-hannover.de}}}%
}
\affil[1]{University of Duisburg-Essen, Germany}
\affil[2]{Leibniz University Hannover, Germany}
\fi

\renewcommand{\headeright}{}
\renewcommand{\undertitle}{}
\renewcommand{\shorttitle}{Learned Compression of Point Cloud Geometry and Attributes in a Single Model through Multimodal Rate-Control}

\hypersetup{
pdftitle={Learned Compression of Point Cloud Geometry and Attributes in a Single Model through Multimodal Rate-Control},
pdfsubject={cs.MM, cs.CV},
pdfauthor={Michael Rudolph, Aron Riemenschneider, Amr Rizk},
pdfkeywords={Virtual Reality, 6DoF, Point Cloud, Variable Rate, Geometry Compression, Attribute Compression},
}

\begin{document}
\maketitle

\begin{abstract}
Point cloud compression is essential to experience volumetric multimedia as it drastically reduces the required streaming data rates.
Point attributes, specifically colors, extend the challenge of lossy compression beyond geometric representation to achieving joint reconstruction of \textit{texture and geometry}.
State-of-the-art methods separate geometry and attributes to compress them individually. 
This comes at a computational cost, requiring an encoder and a decoder for each modality.
Additionally, as attribute compression methods require the same geometry for encoding and decoding, the encoder emulates the decoder-side geometry reconstruction as an input step to project and compress the attributes.

In this work, we propose to learn joint compression of geometry and attributes using a single, adaptive autoencoder model, embedding both modalities into a unified latent space which is then entropy encoded.
Key to the technique is to replace the search for trade-offs between rate, attribute quality and geometry quality, through conditioning the model on the desired qualities of both modalities, bypassing the need for training model  ensembles.
To differentiate important point cloud regions during encoding or to allow view-dependent compression for user-centered streaming, conditioning is pointwise, which allows for local quality and rate variation.
Our evaluation shows comparable performance to state-of-the-art compression methods for geometry and attributes, while reducing complexity compared to related compression methods.
\end{abstract}


\section{Introduction}
\label{sec:intro}
\begin{figure}[t]
    \centering
    \includegraphics[width=.6\columnwidth]{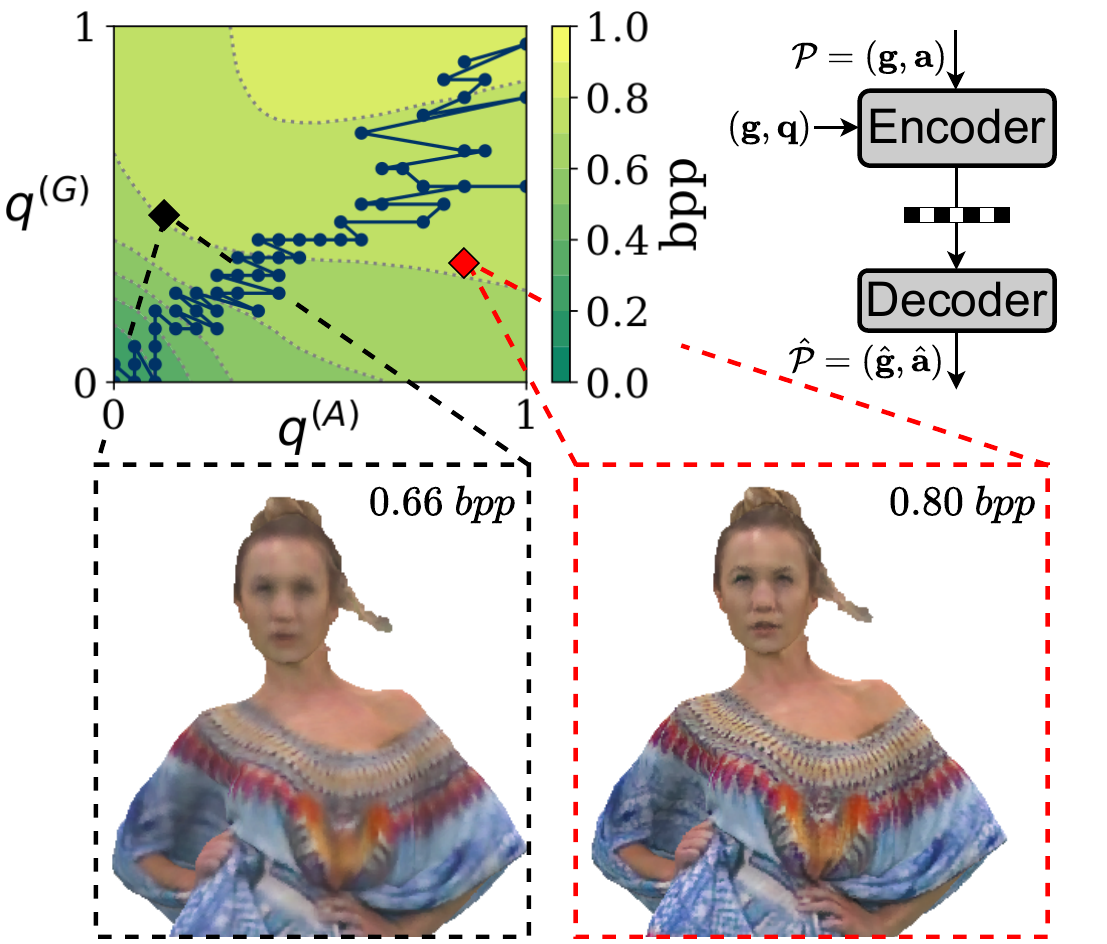}
    \caption{Variable-Rate Coding using a \textit{joint} architecture for geometry and attributes, allows to freely select attribute and geometric quality with\textit{ only a single model}. The figure shows one mapping (path) of attribute and geometry quality configurations $q^{(G/A)}$ that compose a quality (PCQM) Pareto front with the corresponding rates given in bit per point~(bpp).}
    \label{fig:teaser}
\end{figure}

Point clouds are a popular representation for immersive multimedia content \citep{lee2024designing, viola2023vr2gather}, benefiting from low processing overhead compared to other formats for three-dimensional data such as meshes.
The idea of using point clouds as a data structure for streaming dynamical volumetric multimedia experiences over the Internet leads to significant bandwidth requirements, which establishes \textit{the need for efficient compression approaches}.
Standardization proposes solutions for point cloud compression through G-PCC~\citep{gpcc2021}, which uses octree data structures for the geometry and tailored approaches for the attributes like Region Adaptive Hierarchical Transform (RAHT)~\citep{de2016compression}. When considering point cloud sequences, V-PCC~\citep{vpcc2021} utilizes video codecs to compress projections of the point cloud, thus allowing to exploit inter-frame correlations in 2D.

Inspired by and going beyond classical compression methods, Deep Learning has shown promising results for the compression of images \citep{balle2016end, balle2018variational, minnen2018joint}, videos \citep{agustsson2020scale} and point clouds \citep{wang2021lossy,wang2021multiscale, wang2022sparse, quach2019learning, quach2020improved, alexiou2020towards, guarda2019point}. 
Similar to the traditional approach, learned point cloud compression mostly \textit{treats geometry and attribute compression separately}.
A geometry codec handles the compression of the points, while an attribute compression model handles the compression of colors \textit{given the knowledge of the undistorted geometry}.

When combining the individual geometry and attribute codecs, whether learned or traditional, while relying on lossy compression, the reconstructed geometry at the decoder \textit{diverges} from the geometry necessary as a prior for attribute decoding. 
This impacts the quality of the attribute reconstruction.
To address this issue, the geometry is decoded and the attributes are projected on the reconstruction~\citep{gpcc2021, vpcc2021, zhang2023yoga} before attribute encoding, ensuring they are compressed using the same geometric prior which would be available at the receiver side.
We observe, however, that this introduces additional distortions and adds to the computational complexity of the encoding, requiring not only two individual encoding steps but additional geometry decoding and attribute projection, mostly done through costly nearest-neighbor searches, at the sender.

As depicted in Fig.~\ref{fig:teaser}, we propose a joint variable-rate compression method for point clouds in a single, adaptive autoencoder that embeds both modalities into a unified latent space. Figure~\ref{fig:teaser} shows that using this joint model, we can individually gauge the quality of the geometry and the attributes.
Using our single model dispenses with training an ensemble of models to fit a fixed number of individual tradeoffs between the bitrate and the qualities of the geometry and attributes.
The contributions of this paper are:
\begin{itemize}
    \item We formulate the rate-distortion loss for multimodal compression of geometry and attributes to incorporate adaptive weighting of conditional loss components. This allows to train \textit{a single adaptive model} covering a wide range of geometry and attribute qualities and thus rates.
    \item We propose a joint compression model, handling the geometry and attributes in one architecture, avoiding the need for geometry decoding and reprojection on the sender side.
    \item Utilizing conditional encoding, attribute and geometry quality can be modulated independently, \textit{allowing variable rate-encoding} with a single model, as well as, region-of-interest encoding of point clouds.
\end{itemize}

\section{Related Work}
\label{sec:related_work}
Attribute compression is essential for point cloud multimedia applications in contrast to solely compressing geometry, e.g., in LiDAR applications~\citep{He_2022_DensityPreserving}.
Merkuria \textit{et al.} \citep{mekuria2016design} developed a codec utilizing octree partitions and intra-frame prediction while projecting attributes to 2D grids and compressing them with legacy image codecs.
MPEG standards~\citep{schwarz2018emerging} distinguish between static point cloud compression using Geometry-based PCC (G-PCC)~\citep{gpcc2021} and dynamic point cloud compression with Video-based PCC (V-PCC)~\citep{vpcc2021}.
While G-PCC relies on octree partitions for geometry compression and methods like Region Adaptive Hierarchical Transform (RAHT)~\citep{de2016compression} for attributes, V-PCC projects geometry and attribute information into video frames and leverages video codecs for compression. 
Driven by promising results in learned image compression \citep{balle2016end, balle2018variational, minnen2018joint}, learning based solutions for point cloud geometry or attribute compression showed competitive results in both domains.
However, most of these studies~\citep{wang2021lossy, wang2022sparse, wang2023dynamic, quach2019learning, quach2020improved, guarda2019point} do not consider the case where both, geometry and attributes, are subjected to distortions, as it would be the case for a realistic transmission use case. 

We note that early work explored learned compression of attributes and geometry in a single model~\citep{alexiou2020towards}, however, learned compression methods using separate models for both modalities~\citep{zhang2023yoga} were shown to outperform the single model approach.
The latter further boosts performance through wrapping RAHT~\citep{de2016compression} implemented in G-PCC~\citep{gpcc2021} in the attribute branch and applying adaptive filters~\citep{ding2023neural}, which need to be optimized during each encoding, adding to the already large computational cost.

\subsection{Learned Geometry Compression}
Learning-based methods demonstrated competitive performance for point cloud compression, where early work~\citep{quach2019learning, guarda2019point} proposed to use 3D convolutions on voxelized point clouds and leverage learned entropy models~\citep{balle2016end} to optimize a rate-distortion loss.
The reconstruction is formulated as a binary classification problem using the (weighted) Binary Cross Entropy or the Focal Loss~\citep{Lin2017focal} to handle class imbalance due to point clouds sparsity in 3D space.
The authors of~\citep{quach2020improved} employed entropy models from~\citep{minnen2018joint}, allowing to model spatial correlations combined with auto-regressive prediction based on already decoded coefficients.
Utilizing the above reviewed contributions, the authors of~\citep{wang2021lossy} employ a Variational Autoencoder (VAE) architecture based on Voxception-ResNet~\citep{brock2016generative}, allowing to show on-par performance with V-PCC~\citep{vpcc2021} in terms of geometry compression. 
We note, however, as all previously mentioned methods utilize 3D convolutions on sparse data represented in dense data structures, processing the point cloud in cubes is necessary to handle the memory requirements at the cost of limiting throughput and performance.
To overcome this, sparse convolutions~\citep{choy20194d} have been employed in \citep{wang2021multiscale} to process large-scale point clouds in one pass and to exploit global context during hyper-encoding, while a multi-scale classification in each stage of the decoder prunes non-occupied voxels.
Instead of using an encoder-decoder approach, \citep{wang2022sparse} propose to encode the upsampling for each stage individually while relying on group-based decoding to leverage correlations between upsampled voxels. The model in~\citep{wang2022sparse} is then extended for inter-coding of dynamic point cloud sequences~\citep{wang2023dynamic}.

\subsection{Learned Attribute Compression}
Attribute compression is still dominated by traditional approaches, relying on Graph Transform~\citep{zhang2014point} or RAHT~\citep{de2016compression}, with the latter being part of G-PCC.
In the field of learning-based approaches, models using sparse convolutions achieved comparable performance~\citep{wang2022sparse}. 
Combining traditional methods with learning-based approaches, the authors of~\citep{fang20223dac} estimate the probabilities of transform coefficients from RAHT with deep entropy models, further improving compression efficiency.
The attribute compression branch in YOGA~\citep{zhang2023yoga} relies on a thumbnail point cloud compressed at low resolution through RAHT to then predict residuals of the attributes through a conditioned autoencoder similar to~\citep{wang2022sparse}.  
To further increase the quality of the reconstructed attributes, YOGA~\citep{zhang2023yoga} employs adaptive filters~\citep{ding2023neural}, which are optimized during encoding and the parameters are added as side information to the bitstream to be used at the receiver.

\subsection{Variable-Rate Control for Learned Compression}
Numerous compression models for images~\citep{minnen2018joint,balle2018variational} and point clouds~\citep{wang2021lossy,wang2021multiscale, wang2022sparse, quach2019learning, quach2020improved, alexiou2020towards, guarda2019point} based on the non-linear transform coding framework~\citep{balle2016end} achieve remarkable rate-distortion performance.
\textit{A major drawback} of these methods is that they are \textit{conditioned on a fixed tradeoff between compression and reconstruction at training time.}
As a result, an ensemble of models needs to be trained when multiple qualities are required, as it is commonly done in adaptive streaming~\citep{dash2022}. 
Additionally, this further reduces flexibility as encoding configurations need to be \textit{fixed} through hyper-parameters at training time. 

Variable-rate control and progressive coding schemes are promising techniques to freely adapt the rate-distortion tradeoff using a single model for encoding and decoding.
For example, for learned image compression, progressive coding is realized by iteratively refining the quantization grids~\citep{lu2021progressive}. 
Trit-planes~\citep{lee2022dpict} or dead-zone quantizers~\citep{li2022learned} extend this idea to better fit the distribution of quantization errors.
Following a different approach, quantization or reconstruction results are compressed in cascading~\citep{cai2019novel} or recurrent compression~\citep{diao2020drasic, islam2021image,johnston2018improved} models, allowing for progressive coding. 
In contrast to these architectural decisions, training with uniformly sampled tradeoff parameters $\lambda$ while dropping a corresponding number of feature channels in the entropy bottlenecks allows progressive coding through sorting feature channels by importance~\citep{Hojjat_2023_CVPR} and selecting a number of coefficients to decode.

Compared to progressive methods, adaptive rate-control targets conditioning a single model on a rate variable to allow controlling the rate-distortion of the compressed representation.
Early work such as in~\citep{cui2021asymmetric} introduces a gain control layer, multiplying and shifting channels of the feature representation to control the quantization before encoding the transformed coefficients in the entropy bottleneck. 
By incorporating multiple gain control layers in the encoder and decoder, features are scaled hierarchically~\citep{song2021variable, duan2023qarv, cai2022high}. 
Additionally, invertible layers allow increasing the reconstruction quality~\citep{cai2022high}.
YOGA~\citep{zhang2023yoga} adopts these ideas for variable-rate compression with the layers proposed in~\citep{duan2023qarv} using sinusoidal embeddings, but requires an \textit{ensemble of entropy models} as well as \textit{selecting appropriate rate parameters} for the wrapped color compression using G-PCC.

Finally, we note that works such as~\citep{song2021variable} emphasizes regions of interest in images by conditioning compression models on 2D quality maps instead of a single scalar quality parameter.
As the quality map is not available on the decoder side, it is inferred from the compressed representation for decoding.
Extending this idea from image compression to point cloud compression is promising as it allows view-dependent encoding of three-dimensional content as done with traditional codecs~\citep{subramanyam2020user,rudolph2022view}.

\section{Challenge and Approach}
\label{sec:method}

\subsection{Problem Statement}
\label{sec:ProblemStatement}

We consider the problem of compressing a point set $\mathcal{P} = (\mathbf{g}, \mathbf{a})$ containing a set of point \textit{locations} $\mathbf{g}$, denoted geometry, and corresponding \textit{attributes} $\mathbf{a}$. 
Specifically, for the geometry, quantized coordinates $\mathbf{g}=\{ (x_i,y_i,z_i) : x_i,y_i,z_i \in \mathbb{N}_0,x_i,y_i,z_i < R \}$ with maximum value $R\in \mathbb{N}$ are assumed\footnote{This allows using sparse data structures and thus memory and computational efficient convolutional neural networks.}. Note that for real-valued coordinates, a voxelized representation could be easily derived by rounding and partitioning the full 3-dimensional space into blocks with size $R^3$.
We refer to attributes (and features) of a sparse tensor at the geometrical location $u \in \mathbf{g}$ as $\mathbf{a}_u$.

We observe that state-of-the-art compression methods~\citep{gpcc2021,vpcc2021,zhang2023yoga} \textit{handle both modalities, i.e., geometry and attributes, separately.}
Specifically, methods such as~\citep{wang2021lossy, wang2022sparse, wang2023dynamic, quach2019learning, quach2020improved, guarda2019point} or ~\citep{wang2022sparse} solely compress geometry or attributes, respectively. 
Methods that compress both modalities such as~\citep{gpcc2021,vpcc2021,zhang2023yoga} still resort to separating the individual compression procedures for geometry and attributes. 
In essence, they encode the geometry first, which is then decoded to project the attributes to the possibly distorted geometry after reconstruction.
Finally, they compress the projected attributes on the distorted geometry. 
On the decoder side, the geometry is decoded and then used as side-information to allow attribute decoding. 
While this process allows compression of point cloud attributes under geometric distortion, it has a number of disadvantages: i) Encoding and decoding requires a sequence of non-parallelizable steps, resulting in high latency and ii) reprojecting the attributes on distorted geometries at encoding introduces additional distortions before compression.

Now, we aim at training a model for joint attribute and geometry compression.
In a Lagrangian fashion, we seek to minimize the rate $\mathcal{R}$ under the quality constraints on geometry $\mathcal{D}'_G$ and attributes $\mathcal{D}'_A$ given the  loss function 
\begin{equation}
    \mathcal{L} = \mathcal{R} + \lambda_A\mathcal{D}'_A + \lambda_G\mathcal{D}'_G \;,
    \label{eq:lagrange_loss}
\end{equation}
with corresponding deterministic, positive \textit{scalars} $\lambda_A,\lambda_G$.
This can also be understood as finding a tradeoff between geometry and attribute quality at a given rate.
Compared to compression models that handle only a single modality in form of $\mathcal{L} = \mathcal{R} + \lambda \mathcal{D}$, with \textit{fixed} hyperparameter $\lambda$, the loss function above adds another dimension to the number of models required to train, tune and deploy, making ensembles of joint models inadequate for most use cases. 
Hence, a key to solving this challenge is providing a single, adaptive model. This stands in contrast to the image compression literature~\citep{balle2016end, balle2018variational, minnen2018joint} that coined the rate-distortion tradeoff in a bijection sense between the rate $R$ and the quality $D$. In this work, this bijection does not exist as the rate results from defining a tradeoff between geometry and attributes.

A seemingly promising approach is variable rate compression~\citep{song2021variable}, in the sense
\begin{equation}
    \mathcal{L} = \mathcal{R} + \mathcal{D}_A + \mathcal{D}_G \;,
    \label{eq:lagrange_loss_lambda_in_D}
\end{equation}
where we condition a single model for compression at multiple rates, thus only requiring a single training run and reducing the memory requirements at deployment.
In contrast to \eqref{eq:lagrange_loss}, $\mathcal{D}_A,\mathcal{D}_G$ now integrate locality weighting functions  $\boldsymbol{\lambda}_{A}, \boldsymbol{\lambda}_{G}$ instead of scalar quality parameters $\lambda_A,\lambda_G$ allowing to define a location-specific tradeoff between the geometry, attributes and the rate. 
This is especially suited for multimedia point clouds, as it allows to differentiate important regions such as faces, or to offer view-dependent encoding. 
Additionally, instead of searching for tradeoffs between both modalities and the rate in \eqref{eq:lagrange_loss} at training time, e.g. through grid search, attribute and geometric quality can be balanced post-training. 

\subsection{Intermezzo: From Non-Linear Transform Coding to Conditional Transform Coding}
\label{sec:TransformCoding}
\begin{figure}[t]
    \centering
    \subfloat[Transform Coding~\citep{minnen2018joint, balle2018variational}]{\includegraphics[trim={1cm 0.1cm 1cm 0.05cm},clip,width=0.3\columnwidth]{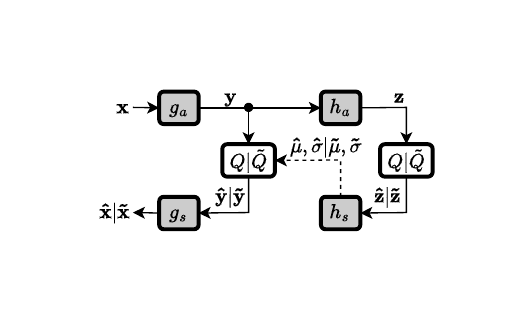} \label{fig:Balle}}
    \hspace{60 pt}
    \subfloat[Conditional extension~\citep{song2021variable}]{\includegraphics[trim={1cm 0.1cm 1cm 0.05cm},clip,width=0.3\columnwidth]{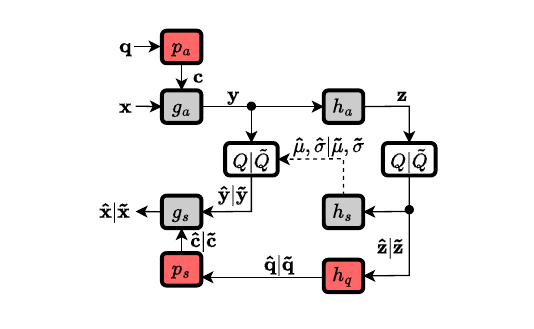} \label{fig:Song}}
    \caption{Operational diagrams of the Mean-Scale Hyperprior model~\citep{balle2018variational, minnen2018joint} and the conditional addition presented by \citep{song2021variable}.}
    \label{fig:TransformCoding}
\end{figure}

Next, we briefly review the non-linear transform coding framework from~\citep{balle2016end} in addition to its extension towards conditional variable rate coding~\citep{song2021variable}.
The following review, also depicted as building blocks in Fig.~\ref{fig:TransformCoding}, considers an arbitrary data vector $\mathbf{x}$. 
In a later section, we show our building blocks where some are based on this framework. 

Figure~\ref{fig:Balle} shows the operational diagram of the mean-scale hyperprior~\citep{balle2018variational, minnen2018joint} where an input data vector $\mathbf{x}$ is non-linearly transformed through an encoder $g_a$. 
The transform output, i.e. the coefficient vector $\mathbf{y}$, is quantized as $\mathbf{\hat{y}} = Q(\mathbf{y})$ and fed to the decoder $g_s$ producing the reconstruction $\mathbf{\hat{x}} = g_s(\mathbf{\hat{y}})$.
Note that the quantized coefficients are usually entropy coded for transmission while the encoder and decoder non-linear transforms $g_a$, and $g_s$ are learned. 

Entropy coding the quantized coefficients $\mathbf{\hat{y}}$ requires an estimate of their distribution. 
Given the observation that the coefficients vary locally in mean and scale, a hyperprior model allows to parameterize a local estimate in form of a Gaussian distribution~\citep{balle2018variational, minnen2018joint}. 
A hyperprior $\mathbf{z}=h_s(\mathbf{y})$ is inferred from the coefficients $\mathbf{y}$ as depicted in  Figure~\ref{fig:Balle}.
The quantized hyperprior $\mathbf{\hat{z}}$ is then entropy coded using a factorized entropy model~\citep{balle2018variational} and used to locally reconstruct the parameters of the Gaussian distributions necessary for encoding/decoding $\mathbf{\hat{y}}$ using the hyperdecoder $h_s$.
Since backpropagation through a quantization operation results in gradients becoming zero almost everywhere, quantization is replaced through additive uniform noise~\citep{balle2016end} for training, i.e. $\mathbf{\tilde{y}} =\mathcal{\tilde{Q}}(\mathbf{y}) = y + \Delta$ with $\Delta$ being an independently and identically distributed uniform random variable, $\Delta \sim \mathcal{U}_{[-0.5, 0.5]}$. In the following, we refer to all variables affected by this quantization proxy through~$\tilde{(\cdot)}$.

Extending this approach to allow variable rate coding, the encoder $g_a$ in Fig.~\ref{fig:Song} is now conditioned on a variable $\mathbf{c}$ which is inferred through the conditional encoder $p_a$ from a quality variable $\mathbf{q}$~\citep{song2021variable}.
This quality variable allows to locally dictate the quality of corresponding elements in the data vector $\mathbf{x}$.
As the quality variable is not available for decoding, a proxy quality variable $\mathbf{\hat{q}}$ is inferred from the hyperprior $\mathbf{\hat{z}}$.
This allows to derive the condition variable $\mathbf{\hat{c}}$ required in the synthesis transform $g_s$.

\subsection{Multimodal Conditioning for Geometry and Attributes}
\label{sec:multimodal}
Next, we show how to achieve variable rate compression of \textit{geometry and attributes} in a single adaptive model through integrating the multimodal rate-distortion loss in \eqref{eq:lagrange_loss_lambda_in_D} into the conditioned transform coding (cf. Sec.~\ref{sec:TransformCoding}). 
To this end, we provide a per-point quality map $\mathbf{q}=[\mathbf{q}^{(G)}, \mathbf{q}^{(A)} ]$ to dictate the target quality individually for both modalities, i.e. geometry and attributes, to the model.
The quality map instantiates the quality variable $\mathbf{q}$ in Sec.~\ref{sec:TransformCoding} residing on the object geometry $\mathbf{g}$. 
During training, each element of the quality map, which is set to $\mathbf{q}^{(A)}_u, \mathbf{q}^{(G)}_u \in [0, 1]$, with $u \in \mathbf{g}$, is mapped to a corresponding weighting map element $\boldsymbol{\lambda}^{(A)}_u, \boldsymbol{\lambda}^{(G)}_u$ through a monotonically increasing function~\citep{song2021variable}.
This allows to individually define the range of the weights in $\boldsymbol{\lambda}^{(A)},\boldsymbol{\lambda}^{(G)}$ and thus the target distortions for the respective modality. 
We use a quadratic weighting function\footnote{We note that exponential functions can be used as in~\citep{song2021variable}, however, experiments show that a root function is more suitable.}, i.e. $T_A: [0, 1] \mapsto {[\lambda_{\text{min}}, \lambda_{\text{max}}]}$ with $T_A(x) = (\lambda^A_{\text{max}} - \lambda^A_{\text{min}}) x^2 + \lambda^A_{\text{min}}$, such that $\boldsymbol{\lambda}^{(A)}_u = T_A(\mathbf{q}^{(A)}_u)$, with $u \in \mathbf{g}$.
The same mapping holds for $\boldsymbol{\lambda}^{(G)}_u$.
The range $[\lambda^{A/G}_{\text{min}}, \lambda^{A/G}_{\text{max}}]$ defines the minimum and maximum values in the form of distinct hyperparameters.
At training time, quality maps $\mathbf{q}$ are either uniformly drawn or supplied as a gradient in a random direction, allowing the model to vary the compression ratio both globally and locally.

\subsubsection{Geometry Loss}
\label{sec:GeometryLoss}
For the geometry reconstruction, we use the Focal Loss~\citep{Lin2017focal}
at every level of the decoder as proposed by~\citep{wang2021multiscale} to allow iterative upsampling of the features and geometry in $K$ steps\footnote{corresponding to $K$ resolutions also denoted as scales throughout this work.}.
In a nutshell, iterative upsampling in $K$ steps is required for constructing the point cloud geometry from the downsampled representation of the encoder. 
More specifically, in each upsampling step $k \in \{0,\dots,K-1\}$, a set of possible point locations $\mathbf{\tilde{d}}^{(k)}$, is generated from the lower scale geometry $\mathbf{\tilde{g}}^{(k+1)}$ s.t. $\mathbf{\tilde{g}}^{(k+1)} \subset \mathbf{\tilde{d}}^{(k)}$~\citep{gwak2020gsdn}\footnote{Later, we omit the scale superscript for $k=0$, i.e. the original resolution $\mathbf{\tilde{g}}^{(0)}:=\mathbf{\tilde{g}}, \mathbf{{g}}^{(0)}:=\mathbf{{g}}$.}.
Finally, an occupancy probability $\mathbf{\tilde{p}}_u^{(k)}$ at location $u$ and scale $k$ is inferred for the possible point locations $\mathbf{\tilde{d}}^{(k)}$.
Incorporating the geometry loss map $\boldsymbol{\lambda}_u^{(G, k)}$ in each scale $k$, the distortion loss $\mathcal{D}_G$ in \eqref{eq:lagrange_loss_lambda_in_D} is
\begin{align}
    &\mathcal{D}_G = - \sum_{k=0}^{K-1} \sum_{u \in \mathbf{\tilde{d}}^{(k)} } \boldsymbol{\lambda}_u^{(G, k)} (1 -\dot{\mathbf{p}}_u^{(k)} ))^{\gamma} \log(\dot{\mathbf{p}}_u^{(k)}) \\
    &\text{ with } \dot{\mathbf{p}}_u^{(k)} = \begin{cases}
        \mathbf{\tilde{p}}_u^{(k)} &\text{if } u \in \mathbf{g}^{(k)} \\
        1 - \mathbf{\tilde{p}}_u^{(k)}  &\text{else}
    \end{cases} \; ,
\end{align}
where $\mathbf{g}^{(k)}$ is the ground truth geometry  at scale $k$ and the hyperparameter $\gamma$ known from the focal loss~\citep{Lin2017focal} reduces the gradients for well-classified samples, allowing the model to emphasize harder samples.
As the loss map $\boldsymbol{\lambda}_u^{(G, k)}$ is only given on ground truth voxel geometry, i.e. $u \in \mathbf{g}$, substitute values $\boldsymbol{\lambda}_u^{(G, k)}$ are derived through downsampling and upsampling using average pooling and transposed average pooling, respectively, resulting on the same base geometry as in $\mathbf{\tilde{d}}^{(k)}$. 

\begin{figure*}
    \centering
    \includegraphics[width=\textwidth]{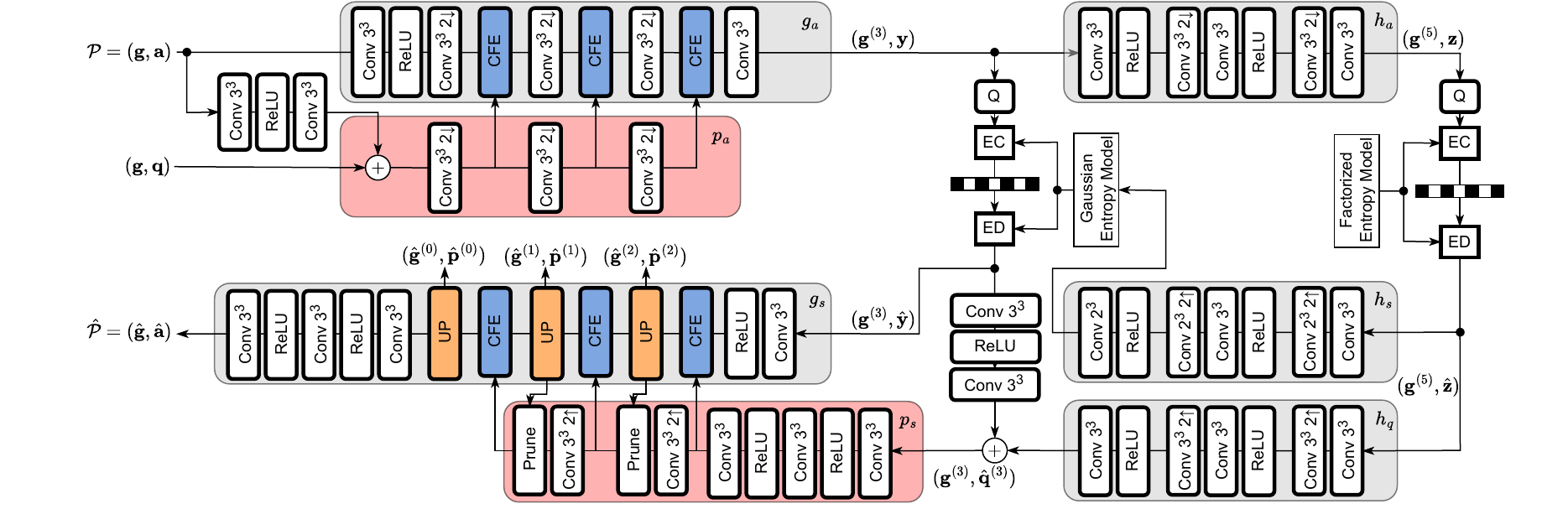}
    \caption{Model Architecture, materializing the operating diagram in Fig.~\ref{fig:Song}. For Convolutions, $b^3$ indicates filters with kernel size $b$ and dimension $3$. The arrows $\uparrow/\downarrow$ symbolize upsampling and downsampling, respectively. A more detailed version of the architecture containing implementation details is given in the supplementary materials.}
    \label{fig:architecture}
\end{figure*}

\subsubsection{Attribute Loss}
\label{sec:AttributeLoss}
The attribute reconstructions is typically cast as an element-wise regression problem~\citep{wang2022sparse, alexiou2020towards}.
Using the $L_2$-norm the attribute component $\mathcal{D}_A$ in  \eqref{eq:lagrange_loss_lambda_in_D} becomes
\begin{equation}
    \mathcal{D}_A = \sum_{u \in \mathbf{g} \cap \tilde{\mathbf{g}}} \boldsymbol{\lambda}_u^{(A)} \lVert \mathbf{a}_u - \mathbf{\tilde{a}}_u \rVert_2^2 \; ,
\end{equation}
where  $\mathbf{a}_u$ denotes the attributes at the geometrical location $u \in \mathbf{g}$ and $\mathbf{\tilde{a}}_u, \tilde{\mathbf{g}}$ are, as discussed in Sect.~\ref{sec:TransformCoding},  affected by the quantization proxy. Recall that the attribute loss map $\boldsymbol{\lambda}_u^{(A)}$ contains individual, i.e. local, weighting to gauge the attributes to target distortion. 

Note that the loss can only be computed on locations $u \in \mathbf{g} \cap \tilde{\mathbf{g}}$, i.e. the intersection of ground truth geometry $\mathbf{g}$ and the reconstruction~$\mathbf{\tilde{g}}$, where we compare the ground truth attributes $\mathbf{a}_u$ with the reconstruction $\tilde{\mathbf{a}}_u$.

\subsubsection{Rate Loss}
The rate loss $\mathcal{R}$ is computed according to~\citep{balle2018variational}, aiming to minimize the cross-entropy as in
\begin{equation}
\begin{aligned}
    \mathcal{R} = &- \sum_{u \in \mathbf{g}^{(K)}}\sum_i \log_2\left(p_{\mathbf{\tilde{y}} |  \mathbf{\tilde{\mu}}, \mathbf{\tilde{\sigma}}}(\mathbf{\tilde{y}}_{u,i} |  \mathbf{\tilde{\mu}}_{u,i}, \mathbf{\tilde{\sigma}}_{u,i})\right)\\
     &- \sum_{u \in \mathbf{g}^{(H)}}\sum_i\log_2\left(p_{\mathbf{\tilde{z} | \phi}}(\mathbf{\tilde{z}}_{u,i}  | \phi^{(i)})\right),
\end{aligned}
\end{equation}
where $\Phi=\{\phi^{(i)}\}_i$ are the learnable parameters of the factorized entropy model~\citep{balle2018variational}. 
We use $\mathbf{\tilde{y}}_{u,i}$ and $\mathbf{\tilde{z}}_{u,i}$ to denote the transformed coefficient of $\mathbf{\tilde{y}}$ and $\mathbf{\tilde{z}}$ at the $i^{\text{th}}$ channel in the feature dimension at the geometry position~$u$, respectively.
In addition, $\mathbf{\tilde{\mu}}_{u,i}, \mathbf{\tilde{\sigma}}_{u,i} $ denote the local estimates of the means and scales that are obtained from the hyperprior $\mathbf{\tilde{z}}$ as in $(\mathbf{\tilde{\mu}}, \mathbf{\tilde{\sigma}}) = h_s(\mathbf{\tilde{z}})$.
Hereby, the hyperprior $\mathbf{\tilde{z}}$ capturing the estimates resides on lower scale geometry $\mathbf{g}^{(H)}$, i.e. $H$ > $K$.

\subsection{Model Architecture}
\label{sec:Architecture}

We propose a sparse neural network architecture as depicted in Fig.~\ref{fig:architecture}, following the framework proposed by \citep{song2021variable} (cf. \ref{fig:Song}).
Recall the notation conventions from Sec.~\ref{sec:TransformCoding}, with $\hat{(\cdot)}$ denoting variables affected by quantization (i.e. \textit{inference }time), and their \textit{training }time counterparts (quantization proxy) given as $\tilde{(\cdot)}$. 

The model receives a point cloud $\mathcal{P} = (\mathbf{g}, \mathbf{a})$.
The encoder $g_a$ downsamples the point cloud $\mathcal{P}$ to the (resolution) scale $k \in \{0,\dots,K-1\}$, defined in Sect.~\ref{sec:GeometryLoss}, in combination with the quality map $\mathbf{q}$ residing on the same geometry $\mathbf{g}$, to obtain a transformed representation $(\mathbf{\hat{g}}^{(K)}, \mathbf{\hat{y}})$.

\begin{figure}
    \centering
    \includegraphics[trim={0.5cm 0.1cm 0cm 0.0cm},clip,width=.7\columnwidth]{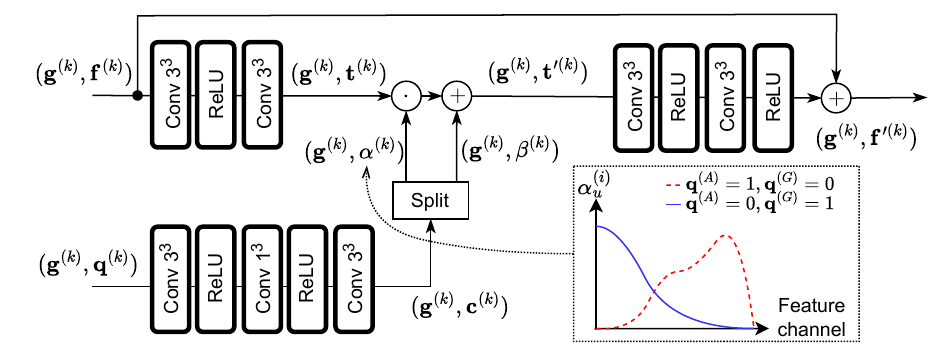}
    \caption{Conditioned feature extraction (CFE) block. A condition map $\mathbf{c}^{(k)} = [\alpha^{(k)}, \beta^{(k)}]$ is derived from the quality map $\mathbf{q}^{(k)}$ at scale $k$ for element-wise scaling and shifting of features according to \eqref{eq:CFE}. Consequently, the module learns a channel-wise weighting of features in $\mathbf{t}^{(k)}$ according to the specified qualities $\mathbf{q}^{(A)}$ and $\mathbf{q}^{(G)}$.}
    \label{fig:CFE}
\end{figure}

\textit{The key to rate adaptive feature transformation} at each scale $k$ is the Conditional Feature Extraction (CFE) block (see Fig.~\ref{fig:CFE}) in the encoder that takes the quality map~$\mathbf{q}$ as input.
On the decoder side, the surrogate\footnote{we call estimates at the decoder side \textit{surrogate}.} quality map $\mathbf{\hat{q}^{(K)}}$ required for adaptive decoding at the lowest scale\footnote{in this work we set the number of resolution scales to $K=3$.} is inferred from the hyperprior $\mathbf{\hat{z}}$ and the transformed coefficients $\mathbf{\hat{y}}$. 
The second ingredient to our method is the Upsampling and Prunining (UP) block during decoding (see Fig.~\ref{fig:UP}) which prunes the geometry at each scale $k$ to maintain the sparsity of the input geometry, thus overcoming memory restrictions~\citep{wang2021multiscale}.

\subsubsection{Conditional Feature Extraction on Geometry and Attribute Importance} 
\label{sec:ConditionBlock}
We employ local conditioning of feature representations through a scale and shift operation, following common procedure to allow variable-rate control in image~\citep{cui2021asymmetric, song2021variable} and point cloud compression~\citep{zhang2023yoga}.
Hereby, features $\mathbf{t}^{(k)}$ residing on the geometry $\mathbf{g}^{(k)}$ at a given scale $k$ are transformed through the \textit{local} scale and shift operation into $\mathbf{t}'^{(k)}$ as
\begin{equation}
    \mathbf{t}_u'^{(k)} = \mathbf{\alpha}_u^{(k)} \mathbf{t}_u^{(k)} + \mathbf{\beta}_u^{(k)} \quad \forall u \in \mathbf{g}^{(k)}.
    \label{eq:CFE}
\end{equation}
using the scales $\mathbf{\alpha}_u^{(k)}$ and shifts $\mathbf{\beta}_u^{(k)}$. 
In contrast to the related work~\citep{zhang2023yoga, song2021variable}, which only predicts element-wise scales $\mathbf{\alpha}_u^{(k)}$ and shifts $\mathbf{\beta}_u^{(k)}$ from a single quality map value $\mathbf{q}^{(k)}_u$ per geometry location $u$, joint adaptation of the geometry and attribute quality requires learning a mapping from \textit{geometry and attribute quality} to the local transformation variables. 

In essence, the feature extraction layer learns the importance of features for optimizing a given task by varying its scale and shift.
This results in broader distributions at the bottleneck when aiming for higher rates, thus reducing quantization errors.
Vice versa, dictating low qualities through $\mathbf{q}^{(k)}$ results in more narrow distributions or in disabling certain feature channels completely through trivial distributions.
Additionally, the additive shifting operation allows to encode the quality map $\mathbf{q}$ into the transformed coefficients~$\mathbf{y}$, embedding information required to infer the surrogate quality map from $\mathbf{\hat{y}}$.

\subsubsection{Upsampling Sparse Features with Geometry Pruning} 
\label{sec:Upsampling}
During encoding, lower scale geometry $\mathbf{g}^{(k+1)}$ is derived from $\mathbf{g}^{(k)}$ through rounding in a dyadic scheme. 
As a result, when upsampling a representation $\mathbf{\hat{g}}^{(k+1)}$ in the decoder, the correct voxels to be occupied in $
\hat{\mathbf{g}}^{(k)}$ are unknown.
Hence, to prevent information loss, voxels are generated at each possible location through a Generative Transposed Convolution~\citep{gwak2020gsdn}, resulting in a dense geometry $\mathbf{\hat{d}}^{(k)}$, which is a superset of $\mathbf{g}^{(k)}$.
While the continuation of decoding with $\mathbf{\hat{d}}^{(k)}$ is conceptually possible, it leads to significant increase in memory and computational requirements.
Geometry compression methods overcome this challenge by performing multiscale classification~\citep{wang2021multiscale, zhang2023yoga}, pruning the upsampled geometry $\mathbf{\hat{d}}^{(k)}$ at each scale $k$ by inferring per-point occupancy probabilities $\mathbf{\hat{p}_u}^{(k)}$ for all $u \in \mathbf{\hat{d}}^{(k)}$.
Note that attribute decoders, such as~\citep{wang2022sparse, zhang2023yoga}, do not face this challenge by \textit{assuming the availability of undistorted geometry} $\mathbf{g}^{(k)}$ from a preceding step, which is in practice not true at the decoder. 

\begin{figure}[t]
    \centering
    \includegraphics[trim={0.5cm 0.3cm 0cm 0cm},clip,width=.65\columnwidth]{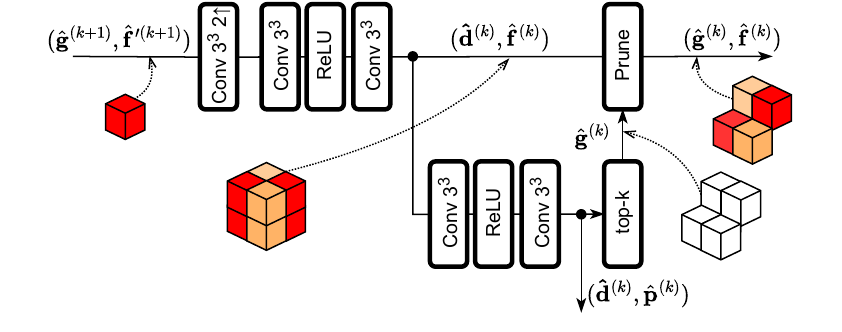}
    \caption{The Upsampling and Pruning (UP) block generates a locally dense geometry {\small$\mathbf{\hat{d}}^{(k)}$} from the lower scale points {\small$\mathbf{\hat{g}}^{(k+1)}$}. Per-point occupancy probabilities {\small $\mathbf{p}^{(k)}$} are predicted from the upsampled and transformed features {\small$\mathbf{f}^{(k)}$}. Finally, the top-k most probable points are selected to prune the dense point cloud {\small$(\mathbf{d}^{(k)}, \mathbf{f}^{(k)})$}, arriving at \small{$(\mathbf{\hat{g}}^{(k)}, \mathbf{\hat{f}}^{(k)})$}.}
    \label{fig:UP}
\end{figure}

Now, when combining both modalities, i.e. geometry and attributes, we need to predict the occupancy and prune the feature representation for the next scale.
The challenge hereby is maintaining an expressive feature representation \textit{after} pruning under potential geometry miss-predictions for attribute reconstruction.
To increase robustness against these miss-predictions, we directly transform the dense, upsampled features, arriving at a dense feature representation, as shown in Fig.~\ref{fig:UP}.  
Then, an occupancy prediction branch infers occupancy probabilities $\mathbf{\hat{p}}^{k}_u$ from the transformed, dense feature representation $\mathbf{f}^{(k)}$.
Finally, top-k selection is performed, keeping the $k$ most probable voxel locations in a pruning step on the transformed, dense features.

\section{Evaluation Results}
\label{sec:Results}

\subsection{Evaluation Setup}
\subsubsection{Implementation}
\label{sec:Implementation}
We implement our approach in PyTorch, using MinkowskiEngine \citep{choy20194d} to leverage sparse convolutions. 
For training, we collect a variety of human point clouds from the UVG dataset~\citep{gautier2023uvg}, select a number of frames for each point cloud sequence and crop them into cubes with edge length $128$, resulting in $18816$ distinct samples.
Opposed to projecting image textures on the samples during training~\citep{wang2022sparse}, we solely subject the attributes to color jitter and randomly rotate each sample. 
For each sample, a random quality map $(\mathbf{g}, \mathbf{q})$ is derived according to Sec.~\ref{sec:multimodal}. 
We train for $200$ epochs using a batch size of $16$.
Adam~\citep{kingma2014adam} is used for optimization with an initial learning rate of $10^{-3}$, which we decrease with factor $0.1$ every $50$ epochs. 
To stabilize training, gradient norm clipping with threshold $1.0$ is integrated.

\subsubsection{Baselines}
\label{sec:baselines}
We compare our method against YOGA~\citep{zhang2023yoga}, which we identify as the current state-of-the-art method for learned compression of geometry \textit{and} attributes. 
Additionally, we use the tmc2v24 test model for V-PCC~\citep{vpcc2021}, using all-intra test conditions and the tmc13v23 implementation of G-PCC~\citep{gpcc2021} employing octree compression for geometry and RAHT~\citep{de2016compression} for attributes. For both methods, we select the first $4$ rate configurations specified in the respective test models, as we find that they result in a similar range of rates as our model.

\subsubsection{Evaluation Metrics}
\label{sec:evalProtocol}
We measure the bitrate in bits per point (bpp), using the cardinality $|\mathbf{g}|$ as divisor.
Geometric quality is assessed using the D1-PSNR~\citep{tian2017geometric} metric, computing the mean distances between points of the tested geometry $\mathbf{g}$ and their nearest neighbor in the reference $\hat{\mathbf{g}}$. 
Utilizing the derived neighbor association, the Y-PSNR and YUV-PSNR metrics~\citep{mekuria2017performance} are computed to assess the attribute quality, with the former only using the luminance channel and the later referring to the weighted average over all channels using weights $\left(\frac{6}{8}, \frac{1}{8}, \frac{1}{8} \right)$, following evaluation protocols in related literature~\citep{wang2022sparse, zhang2023yoga}. 
We consistently present symmetric metrics, i.e. computation is performed for swapped source and reference while reporting the stricter result.
Finally, PCQM~\citep{meynet2020pcqm} is used as a joint metric, incorporating geometric \textit{and} attribute features. 
Note that for all presented results, the reported bitrates correspond to the compression of geometry and attributes together.

We select $4$ frames each from the 8iVFBv2~\citep{upperBodies20178i} and MVUB~\citep{loop2016microsoft} dataset for testing, which are specified in the supplementary material. 
If not specified otherwise, uniform quality maps $\mathbf{q}^{(A)}$ and~$\mathbf{q}^{(G)}$ are used in our model to allow fair comparison.

\begin{figure}
    \centering
    
    \subfloat[\textit{longdress}]{\includegraphics[trim={0cm 0cm 0cm 0cm},clip,width=0.24\textwidth]{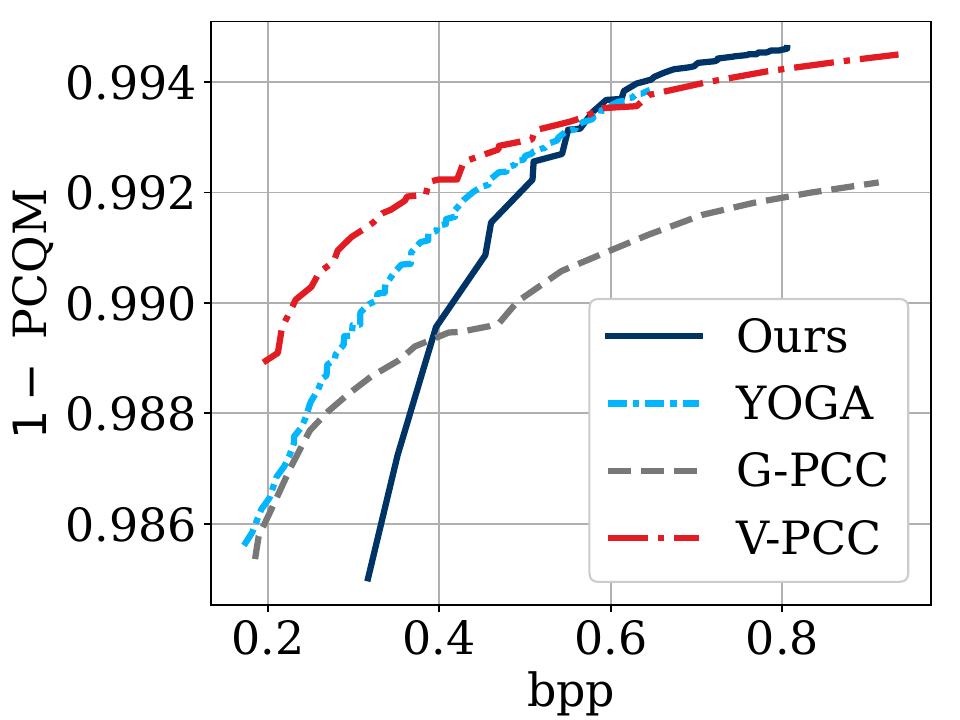} \label{fig:Longdress}}
    \subfloat[\textit{redandblack}]{\includegraphics[trim={0cm 0cm 0cm 0cm},clip,width=0.24\textwidth]{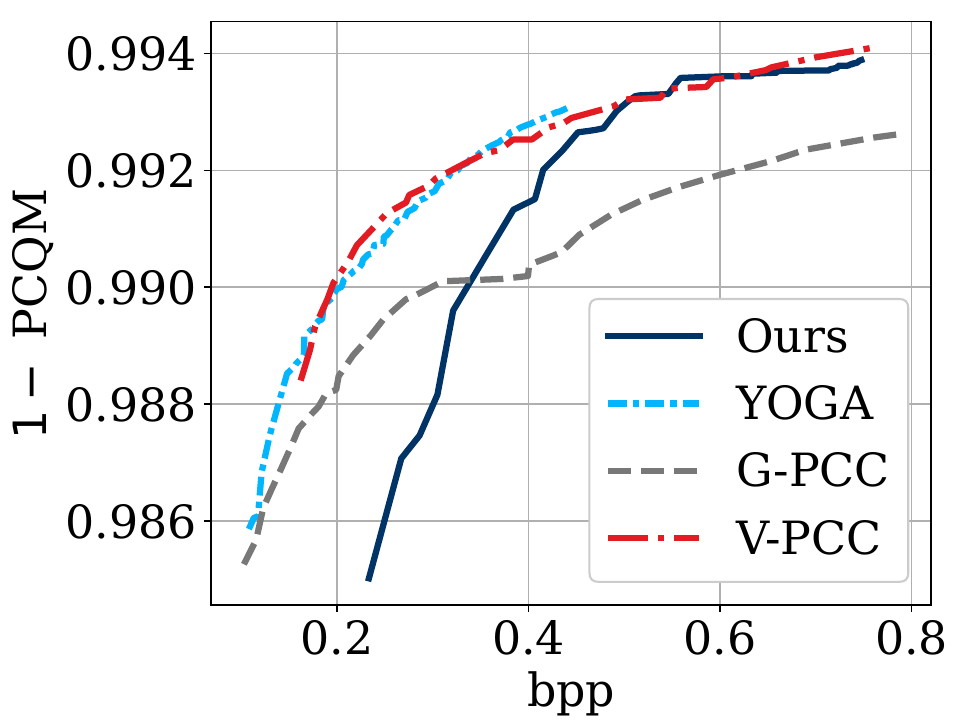} \label{fig:Redandblack}} 
    \subfloat[\textit{soldier}]{\includegraphics[trim={0cm 0cm 0cm 0cm},clip,width=0.24\textwidth]{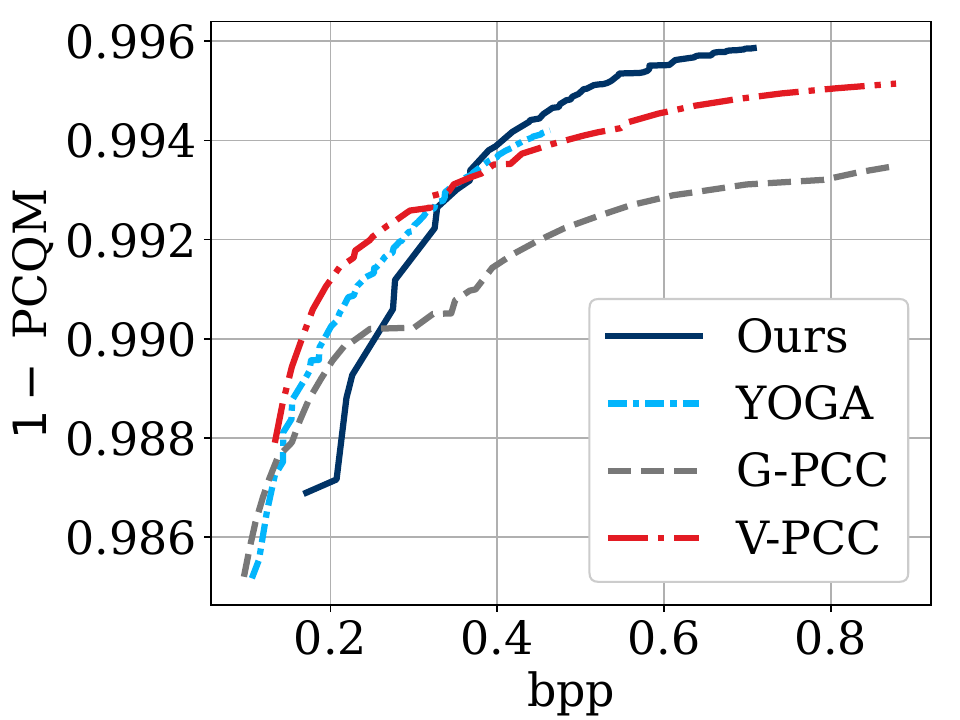} \label{fig:Soldier}}
    \subfloat[\textit{loot}]{\includegraphics[trim={0cm 0cm 0cm 0cm},clip,width=0.24\textwidth]{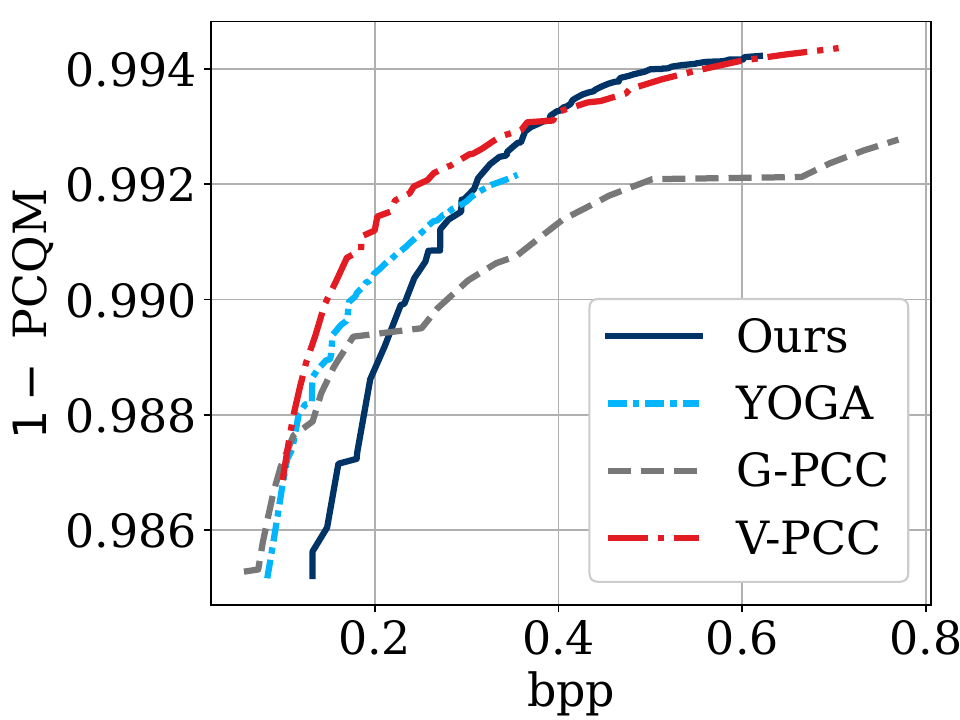} \label{fig:Loot}} \\
    
    \subfloat[\textit{andrew9}]{\includegraphics[trim={0cm 0cm 0cm 0cm},clip,width=0.24\textwidth]{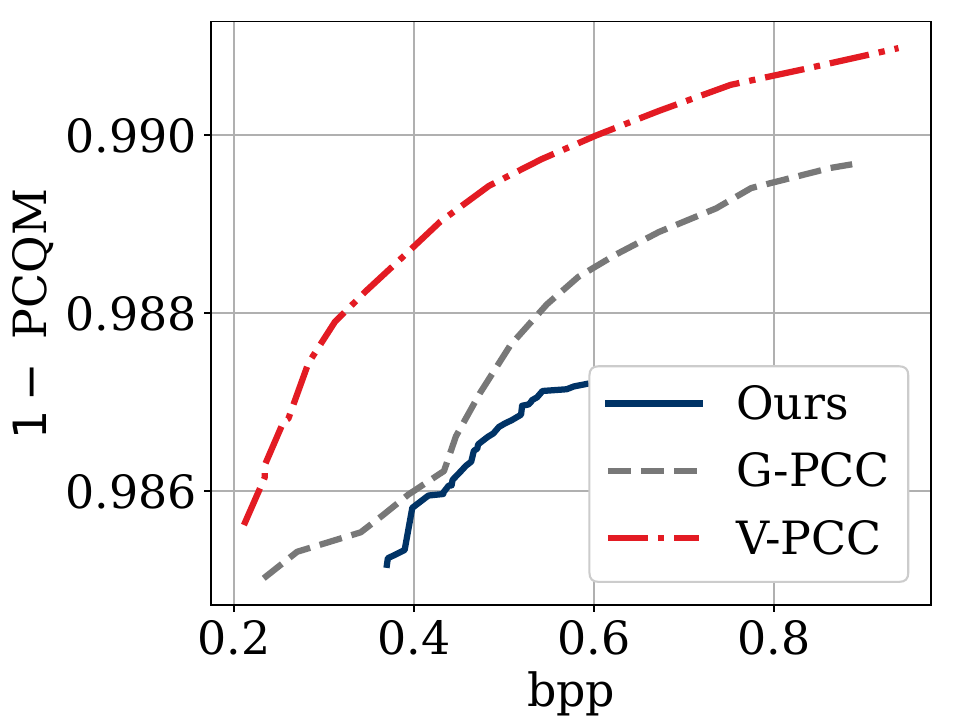}} \label{fig:andrew}
    \subfloat[\textit{phil9}]{\includegraphics[trim={0cm 0cm 0cm 0cm},clip,width=0.24\textwidth]{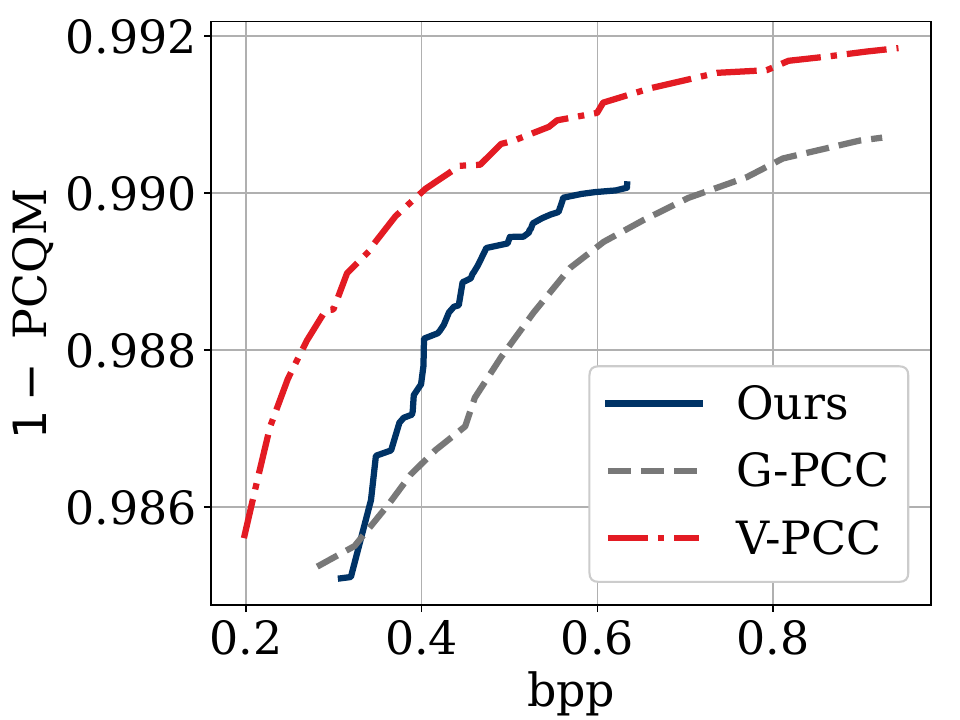}} \label{fig:phil}
    \subfloat[\textit{david9}]{\includegraphics[trim={0cm 0cm 0cm 0cm},clip,width=0.24\textwidth]{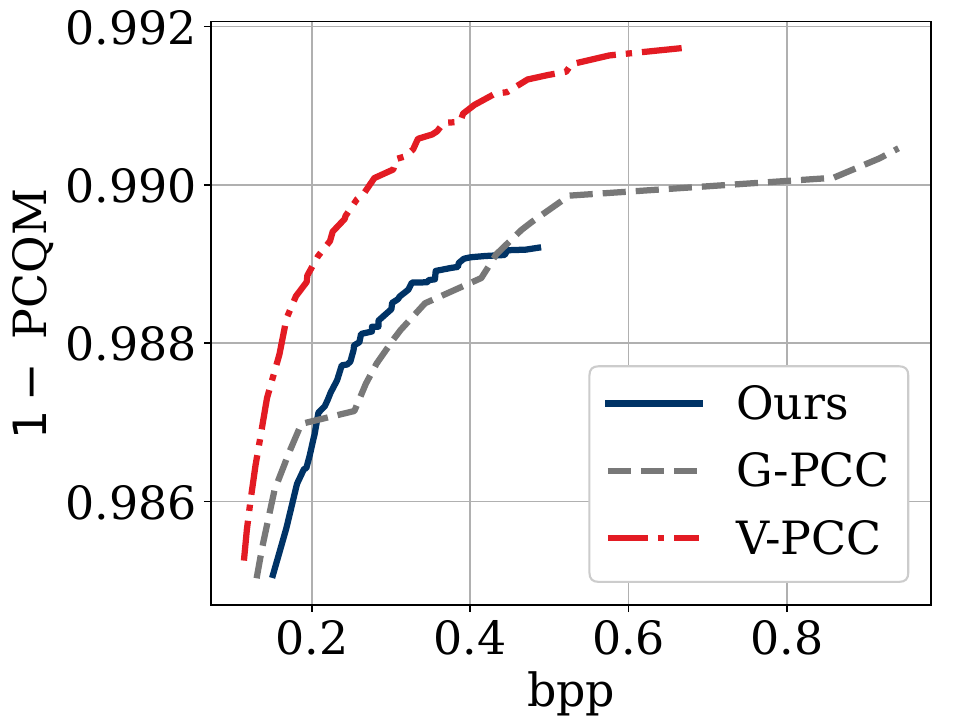}} \label{fig:david}
    \subfloat[\textit{sarah9}]{\includegraphics[trim={0cm 0cm 0cm 0cm},clip,width=0.24\textwidth]{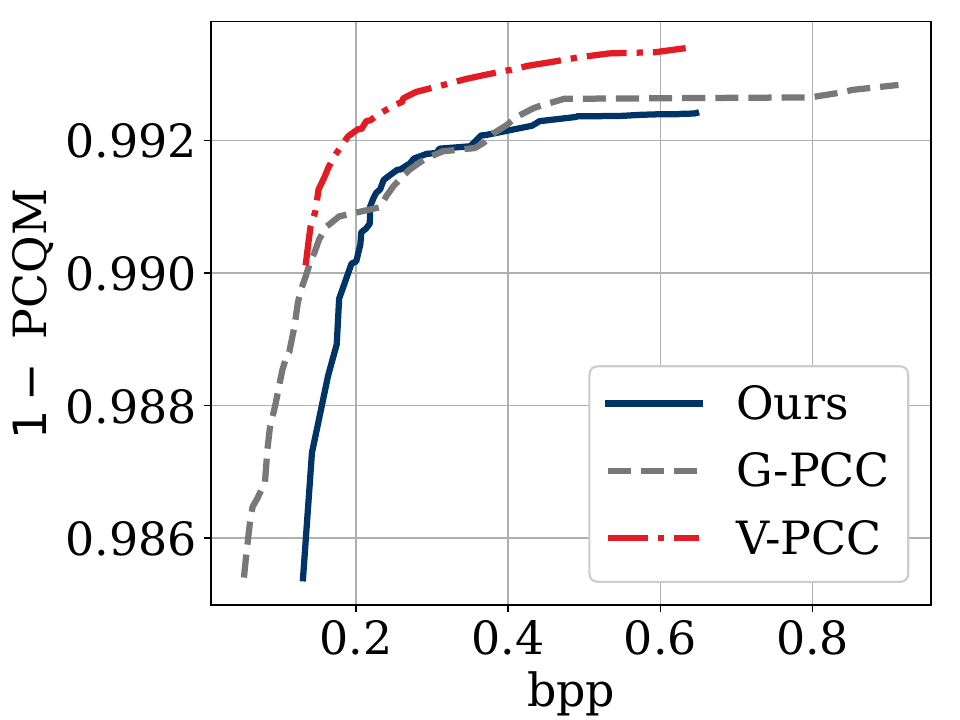}} \label{fig:sarah}
    \caption{Pareto fronts for the PCQM-Rate tradeoff, optimized for the given point cloud.}
    \label{fig:Pareto}
\end{figure}

\begin{figure}
    \centering
    
    \subfloat{\includegraphics[trim={0cm 0cm 0cm 0cm},clip,width=0.33\columnwidth]{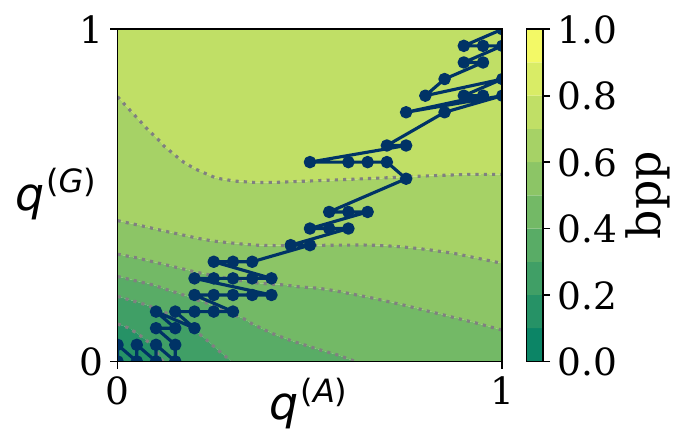} \label{fig:bpp_rab}}
    \subfloat{\includegraphics[trim={0cm 0cm 0cm 0cm},clip,width=0.32\columnwidth]{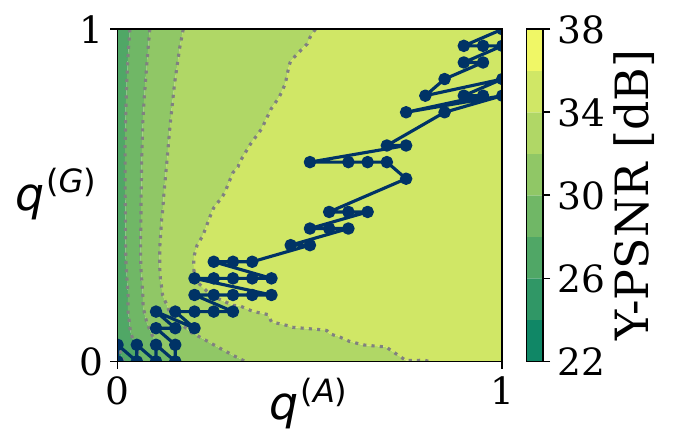} \label{fig:ypsnr_rab}} 
    \subfloat{\includegraphics[trim={0cm 0cm 0cm 0cm},clip,width=0.32\columnwidth]{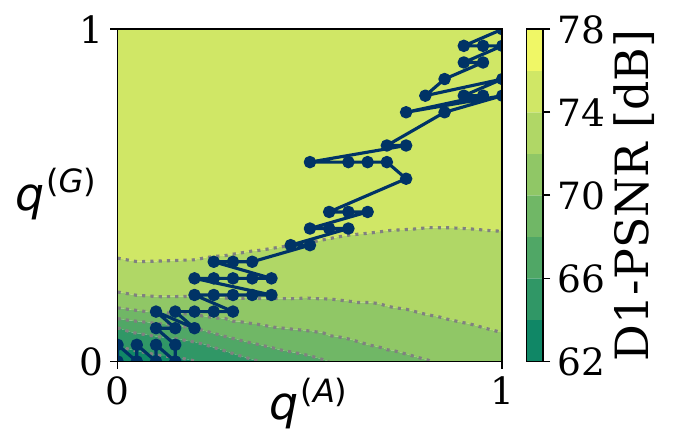} \label{fig:p2ppsnr_rab}}\\

    \setcounter{subfigure}{0}
    \subfloat[bpp]{\includegraphics[trim={0cm 0cm 0cm 0cm},clip,width=0.33\columnwidth]{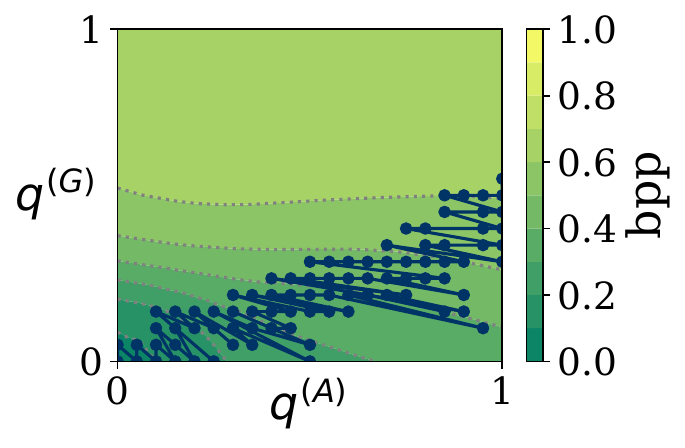} \label{fig:bpp_loot}}
    \subfloat[Y-PSNR]{\includegraphics[trim={0cm 0cm 0cm 0cm},clip,width=0.32\columnwidth]{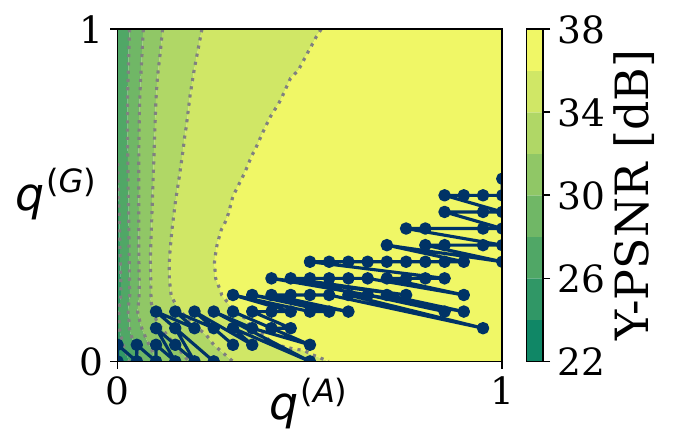} \label{fig:ypsnr_loot}} 
    \subfloat[D1-PSNR]{\includegraphics[trim={0cm 0cm 0cm 0cm},clip,width=0.32\columnwidth]{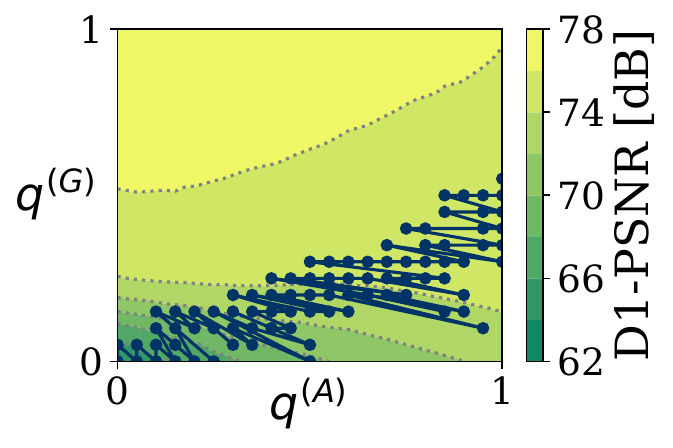} \label{fig:p2ppsnr_loot}}
    \caption{Mapping of attribute and geometry quality configurations $\mathbf{q}^{(A)},\mathbf{q}^{(G)}$ to the metrics bit per point (bpp), Y-PSNR and D1-PSNR for \textit{redandblack} (top) and \textit{loot} (bottom). The paths shows configuration pairs $\mathbf{q}^{(A)},\mathbf{q}^{(G)}$ that stem from the Pareto fronts in Fig.~\ref{fig:Pareto}.}
    \label{fig:Pareto_map}
\end{figure}

\subsection{Perceptual Quality Optimized Pareto Front}
\label{sec:pareto}
Following the evaluation protocol specified in~\citep{zhang2023yoga}, we traverse the cross product of quality configurations for our approach with step size $0.05$, resulting in $400$ configurations for the point clouds in 8iVFBv2~\citep{upperBodies20178i}. 
Then, the Pareto front for the tradeoff between PCQM~\citep{meynet2020pcqm} and rate is determined and depicted in Fig.~\ref{fig:Pareto}. 
Two examples of the chosen configuration pairs can be found in Fig.~\ref{fig:Pareto_map}, which shows the pairs of attribute quality $q^{(A)}$ and geometric quality $q^{(G)}$ contributing to the Pareto front as a path over all possible combinations. 
The underlying contour in each plot shows the bitrate, attribute quality and geometric quality, resulting from all possible combinations.
Similarly, Pareto fronts for V-PCC~\citep{vpcc2021} and G-PCC~\citep{gpcc2021} are determined by traversing the full range of attribute QPs for each geometry setting.
We also compare against the results reported in the original paper of YOGA~\cite{zhang2023yoga} for the 8iVFBv2~\cite{upperBodies20178i} dataset.

From Fig.~\ref{fig:Pareto} we find that our model shows improved PCQM performance compared to all baselines for medium to high rates, indicating comparable or better perceived quality of the reconstructions at this compression ratio. 
However, when aiming for lower rates, the quality deteriorates faster compared to related methods.
We attribute this to the fact that our approach is the only method \textit{solely} relying on learned compression for the point clouds, while, e.g. YOGA~\citep{zhang2023yoga} not only inherits low rate compression capabilities by compressing a thumbnail attribute representation using RAHT~\citep{de2016compression} but also employs adaptive filters~\citep{ding2023neural} which are optimized during encoding on the specific content to counteract quality degradation.

We note that the shown results present optimized configurations for the respective point clouds.
This is also apparent when comparing the selected configuration pairs for the point clouds \textit{redandblack} and \textit{loot} contributing to the Pareto fronts in Fig.~\ref{fig:Pareto}, which are visualized as paths in Fig.~\ref{fig:Pareto_map}.
Depending on the individual attribute and geometric properties of the content, we notice that the paths and thus the chosen set of configurations to arrive at the Pareto fronts, differ significantly.
Recall that this evaluation required traversing a large number of possible configurations to select the best tradeoff points to arrive at the pareto fronts and is thus not practical.
Nevertheless, this evaluation provides insights into the capabilities of the tested compression methods.

\begin{figure*}
    \centering
    \subfloat{\includegraphics[trim={1cm 0cm 0cm 0cm},clip,width=0.242\textwidth]{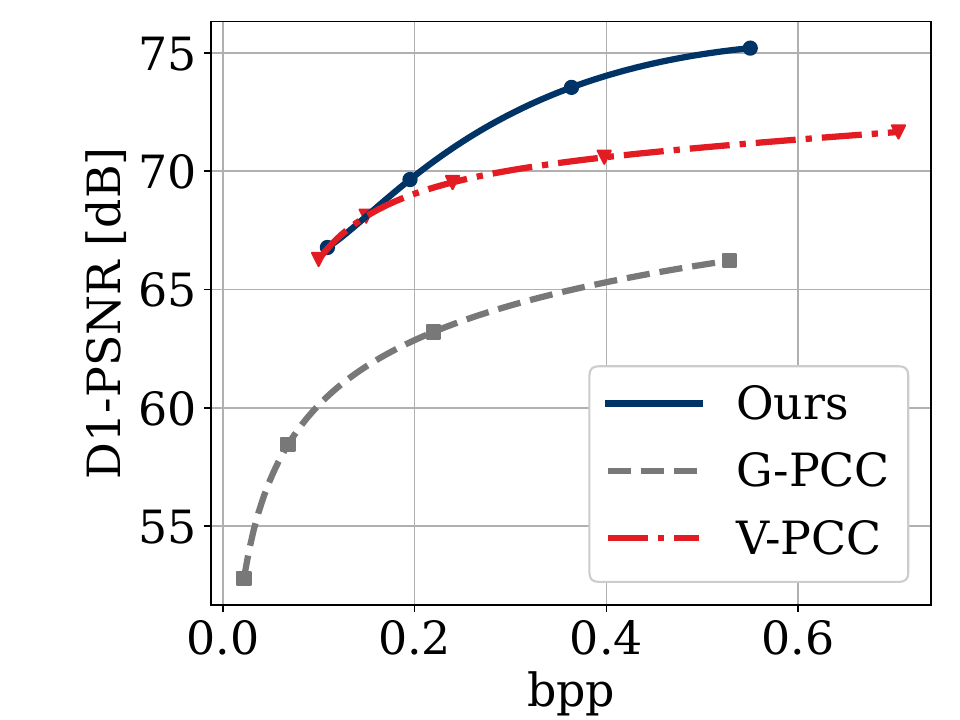} \label{fig:sym_p2p_psnr_loot}}
    \subfloat{\includegraphics[trim={1cm 0cm 0cm 0cm},clip,width=0.242\textwidth]{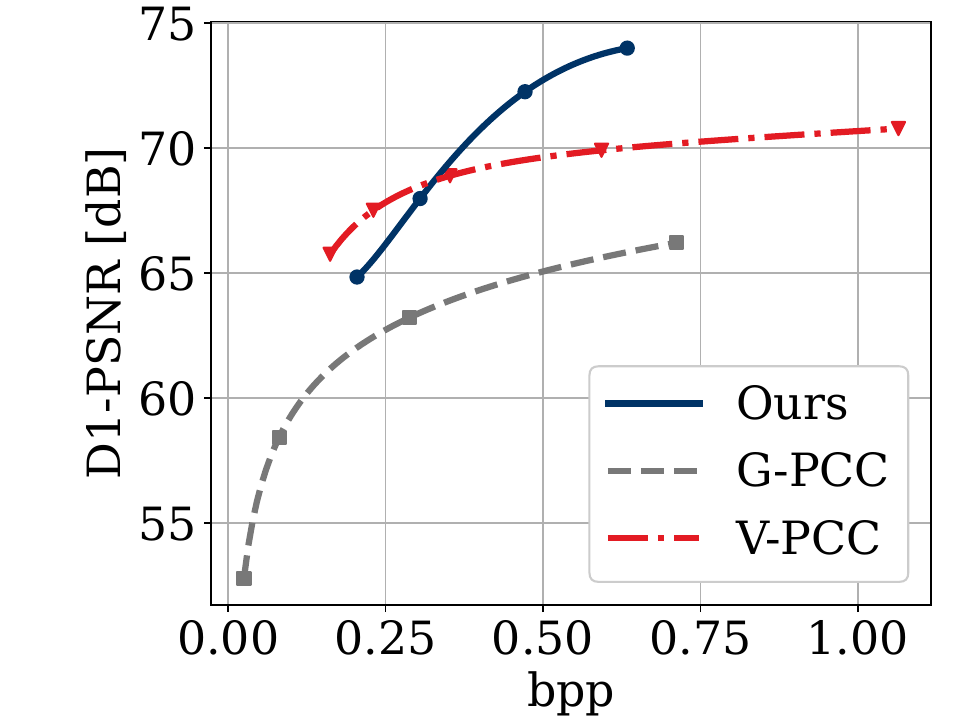}} \label{fig:sym_p2p_psnr_redandblack} 
    \subfloat{\includegraphics[trim={1cm 0cm 0cm 0cm},clip,width=0.242\textwidth]{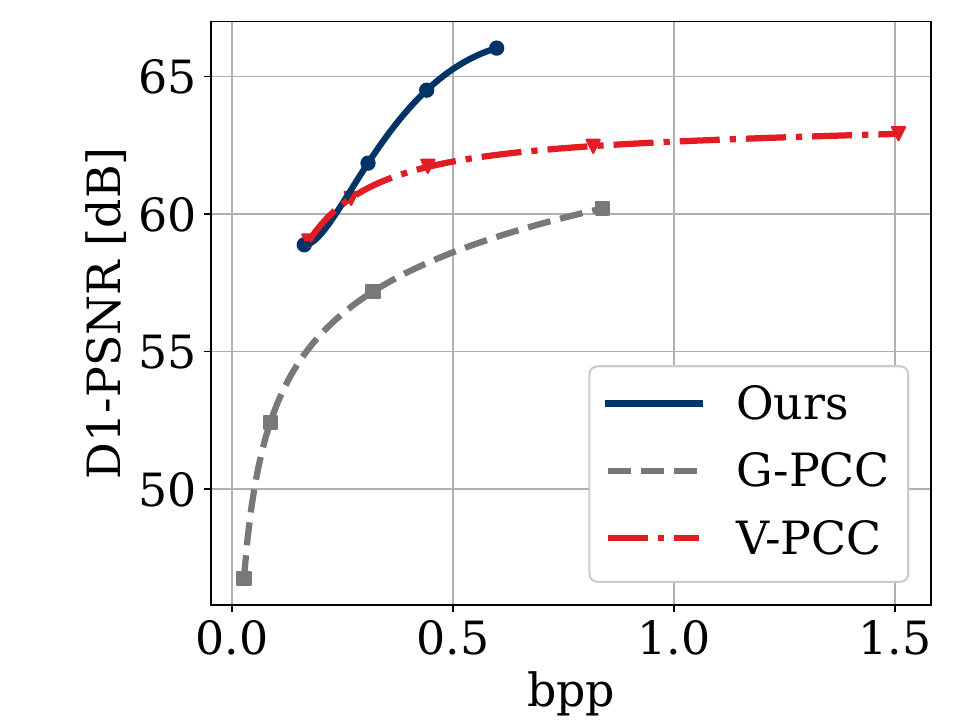} \label{fig:sym_p2p_psnr_phil9}}
    \subfloat{\includegraphics[trim={1cm 0cm 0cm 0cm},clip,width=0.242\textwidth]{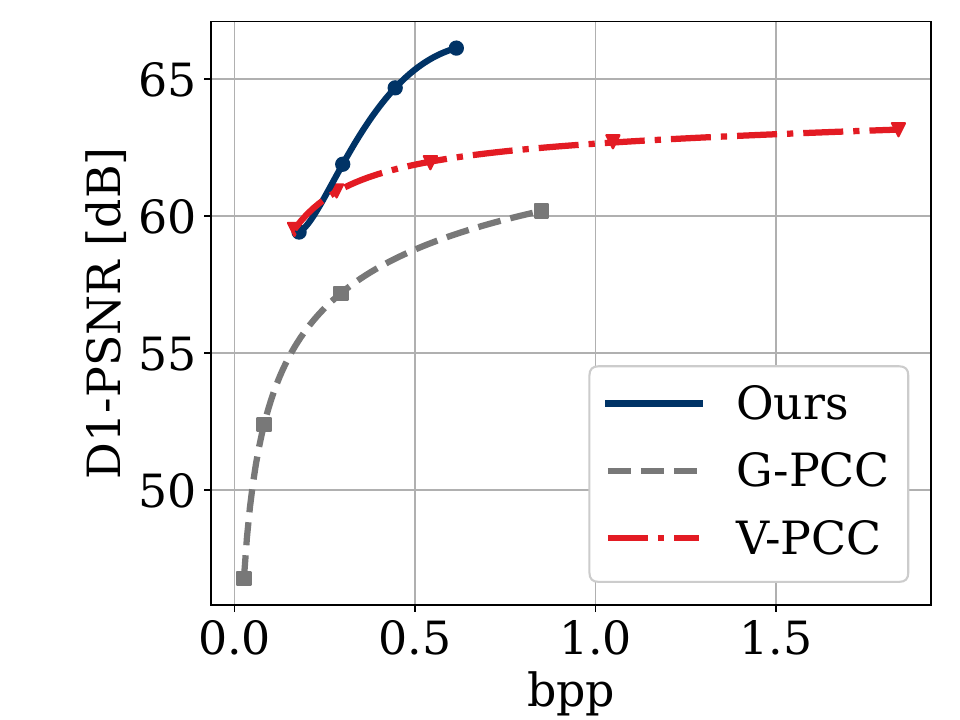} \label{fig:sym_p2p_psnr_andrew9}}
    \\
    \setcounter{subfigure}{0}
    \subfloat[\textit{loot}]{\includegraphics[trim={1cm 0cm 0cm 0cm},clip,width=0.242\textwidth]{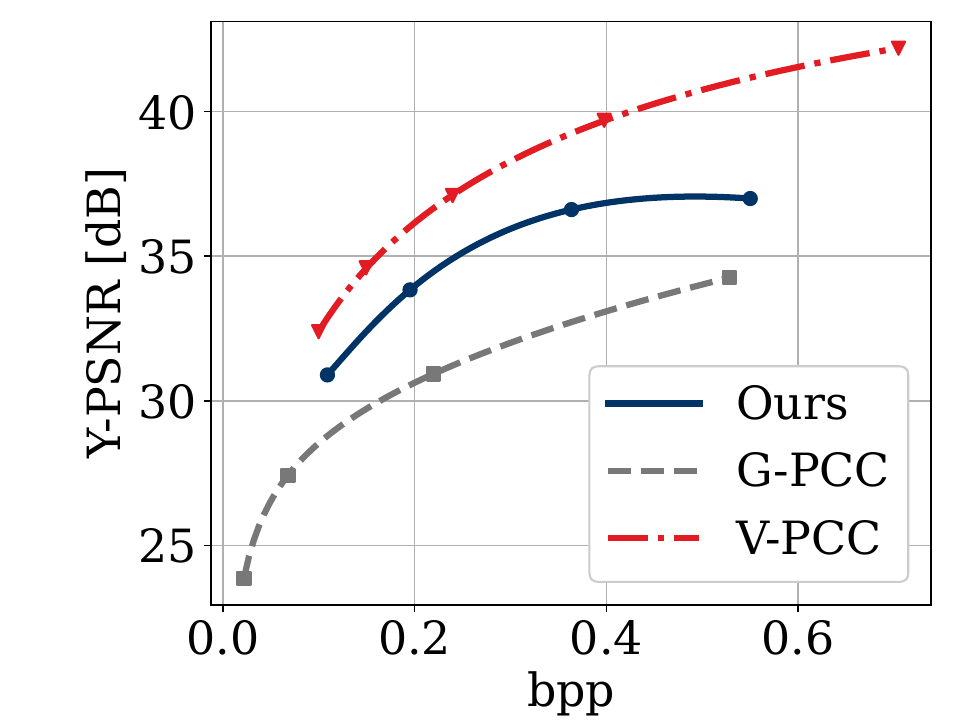} \label{fig:sym_y_psnr_loot}}
    \subfloat[\textit{redandblack}]{\includegraphics[trim={1cm 0cm 0cm 0cm},clip,width=0.242\textwidth]{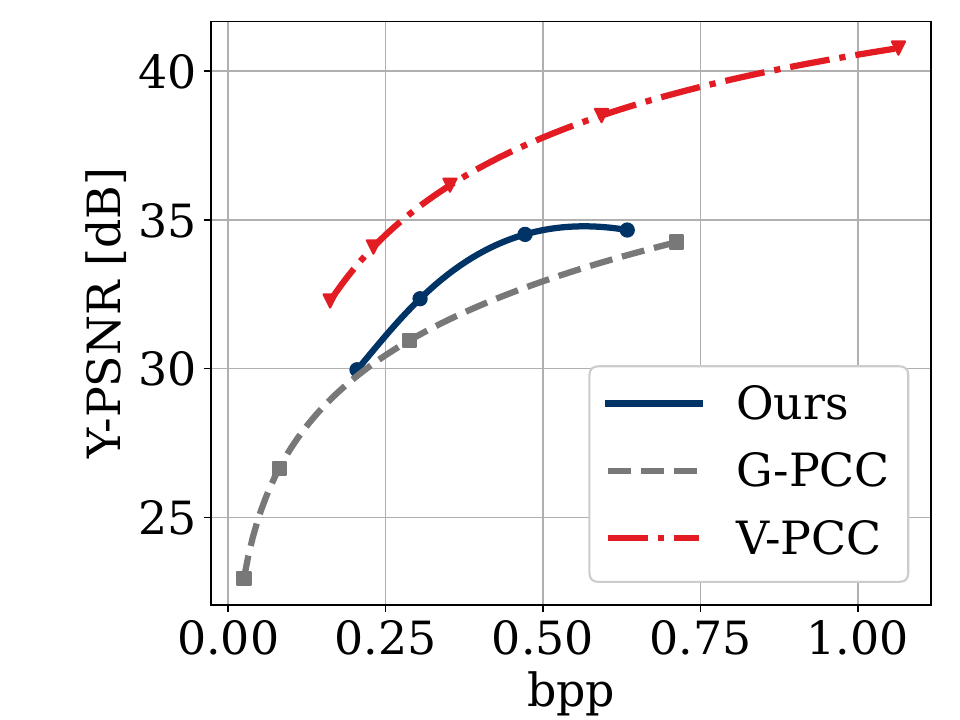} \label{fig:sym_y_psnr_redandblack}} 
    \subfloat[\textit{phil9}]{\includegraphics[trim={1cm 0cm 0cm 0cm},clip,width=0.242\textwidth]{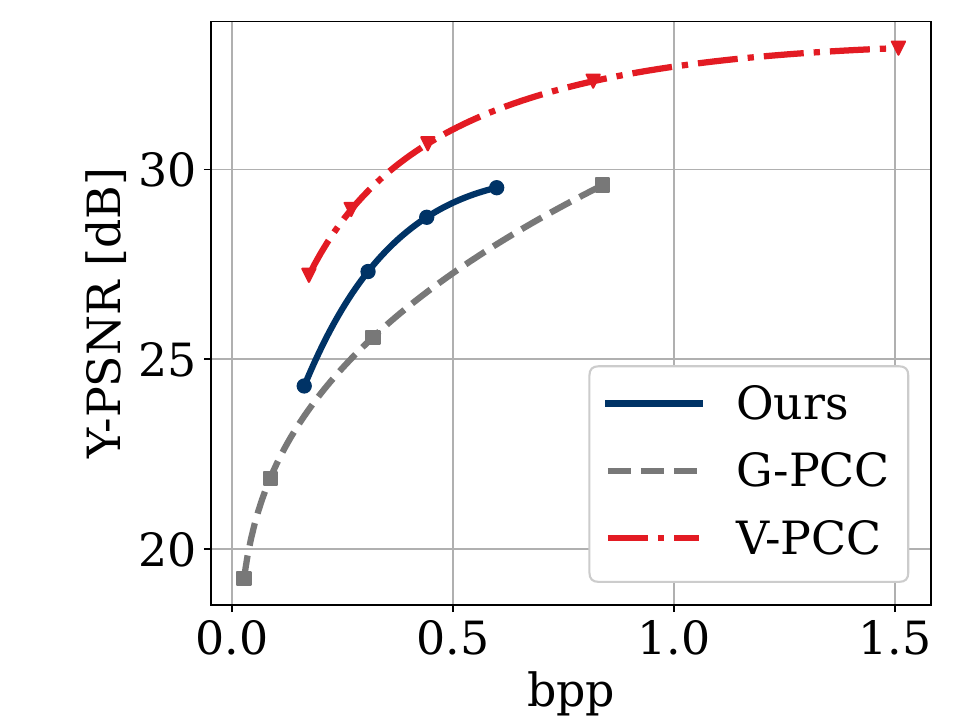} \label{fig:sym_y_psnr_phil9}}
    \subfloat[\textit{andrew9}]{\includegraphics[trim={1cm 0cm 0cm 0cm},clip,width=0.242\textwidth]{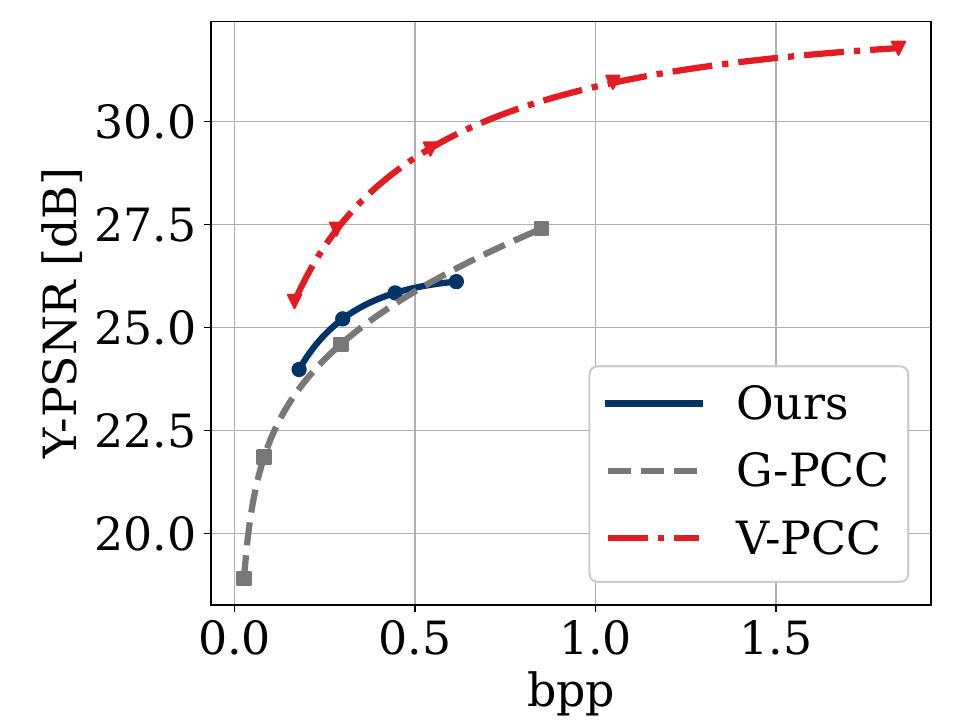} \label{fig:sym_y_psnr_andrew9}}
    \caption{Comparison of rate-distortion for geometry (D1-PSNR, top row) and attributes (Y-PSNR, bottom row) for selected point clouds from 8iVFBv2~\citep{upperBodies20178i} and MVUB~\citep{loop2016microsoft}.}
    \label{fig:rd_results}
\end{figure*}

\subsection{Fixed-Configuration Encoding Performance}
\label{sec:RateDistortion}

Now, we set the conditioning of our model to a fixed set of configurations, i.e. $(\mathbf{q}^{(G)}_u, \mathbf{q}^{(A)}_u) \in \{(0.05, 0.1), (0.1, 0.2),  (0.2, 0.4), \allowbreak(0.4, 0.8)\}$ $\forall u \in \mathbf{g}$, independent of the input point cloud.
This allows to compare against G-PCC~\citep{gpcc2021} and V-PCC~\citep{vpcc2021} using their respective standard configurations. 
We exclude YOGA~\citep{zhang2023yoga}, as such a set of configurations is not publicly available.

Selected rate-distortion curves are presented in Fig.~\ref{fig:rd_results} and corresponding Bj{\o}nte\-gaard-Deltas to G-PCC~\citep{gpcc2021} are shown in Tab.~\ref{tab:rd_table}. 
We provide further rate-distortion curves in the supplementary material.
Note that our approach allows for significantly higher geometric quality according to D1-PSNR, thus small overlap of the resulting rate-distortion curves in the y-axis. This renders the Bj{\o}nte\-gaard-rate Delta $\Delta_r$ for the D1-PSNR metric less expressive on a subset of the tested point clouds.

Over all tested point clouds, our approach shows significantly higher geometric quality compared to G-PCC~\citep{gpcc2021} and V-PCC~\citep{vpcc2021}. 
The corresponding attribute rate-distortion curves according to the Y-PSNR metric show notable gains over G-PCC, but underperform when compared against V-PCC. 
Considering attribute quality according to the YUV-PSNR metric, Tab.~\ref{tab:rd_table} shows similar results, with the exception of sequences with complex textures, i.e \textit{longdress}, \textit{redandblack} and \textit{andrew9}.
For the Bj{\o}nte\-gaard-rate metrics $\Delta_r$ and $\Delta_D$ of the joint metric PCQM~\citep{meynet2020pcqm}, we see high potential for rate-savings in the fixed configuration range compared to G-PCC~\citep{gpcc2021}.

\begin{table}[]
    \centering
    \caption{BD-Results for our approach compared to G-PCC~\citep{gpcc2021}.}
    \begin{tabular}{l|rr|rr|rr|rr}
\toprule
\multicolumn{1}{l|}{} &
  \multicolumn{2}{c|}{D1-PSNR} &
  \multicolumn{2}{c|}{Y-PSNR} &
  \multicolumn{2}{c|}{YUV-PSNR} &
  \multicolumn{2}{c}{$1-$PCQM} \\
\multicolumn{1}{l|}{Point Cloud} &
  \multicolumn{1}{r}{$\Delta_{r} \!\downarrow$} &
  \multicolumn{1}{r|}{$\Delta_{D}\! \uparrow$} &
  \multicolumn{1}{r}{$\Delta_{r} \!\downarrow$} &
  \multicolumn{1}{r|}{$\Delta_{D}\! \uparrow$} &
  \multicolumn{1}{r}{$\Delta_{r} \!\downarrow$} &
  \multicolumn{1}{r|}{$\Delta_{D}\! \uparrow$} &
  \multicolumn{1}{r}{$\Delta_{r} \!\downarrow$} &
  $\Delta_{D} \uparrow$ \\
\midrule
  \textit{longdress} &
    -82.5\%&
    7.49 dB&
    -3.6\%&
    1.07 dB&
    22.6\%&
    0.10 dB&
    -13.7\%&
    0.0016\\
  \textit{soldier} &
    -82.6\%&
    6.78 dB&
    -39.8\%&
    2.94 dB&
    -24.3\%&
    1.70 dB&
    -43.0\%&
    0.0063\\
  \textit{loot} &
    -82.5\%&
    7.49 dB&
    -45.8\%&
    3.28 dB&
    -37.4\%&
    1.89 dB&
    -51.8\%&
    0.0060\\
  \textit{redandblack} &
    -60.4\%&
    5.63 dB&
    -19.5\%&
    1.24 dB&
    -3.1\%&
    0.59 dB&
    -17.6\%&
    0.0023\\
\midrule
  \textit{andrew9} &
    -72.8\%&
    5.16 dB&
    -16.3\%&
    0.38 dB&
    0.4\%&
    -0.02 dB&
    -19.3\%&
    0.0030\\
  \textit{david9} &
    79.0\%&
    6.91 dB&
    -48.2\%&
    2.70 dB&
    -42.6\%&
    2.20 dB&
    -52.3\%&
    0.0069\\
  \textit{phil9} &
    -70.7\%&
    5.05 dB&
    -34.2\%&
    1.63 dB&
    -26.6\%&
    1.05 dB&
    -25.1\%&
    0.0071\\
  \textit{sarah9} &
    -71.1\%&
    6.68 dB&
    -36.1\%&
    2.90 dB&
    -33.8\%&
    2.36 dB&
    -13.1\%&
    0.0039\\ 
\bottomrule
  
\end{tabular}
    \label{tab:rd_table}
\end{table}

\begin{figure}
    \centering
    \setcounter{subfigure}{0}
    \subfloat[GT  ]{\includegraphics[trim={29cm 20cm 27.4cm 5cm},clip,width=0.25\columnwidth]{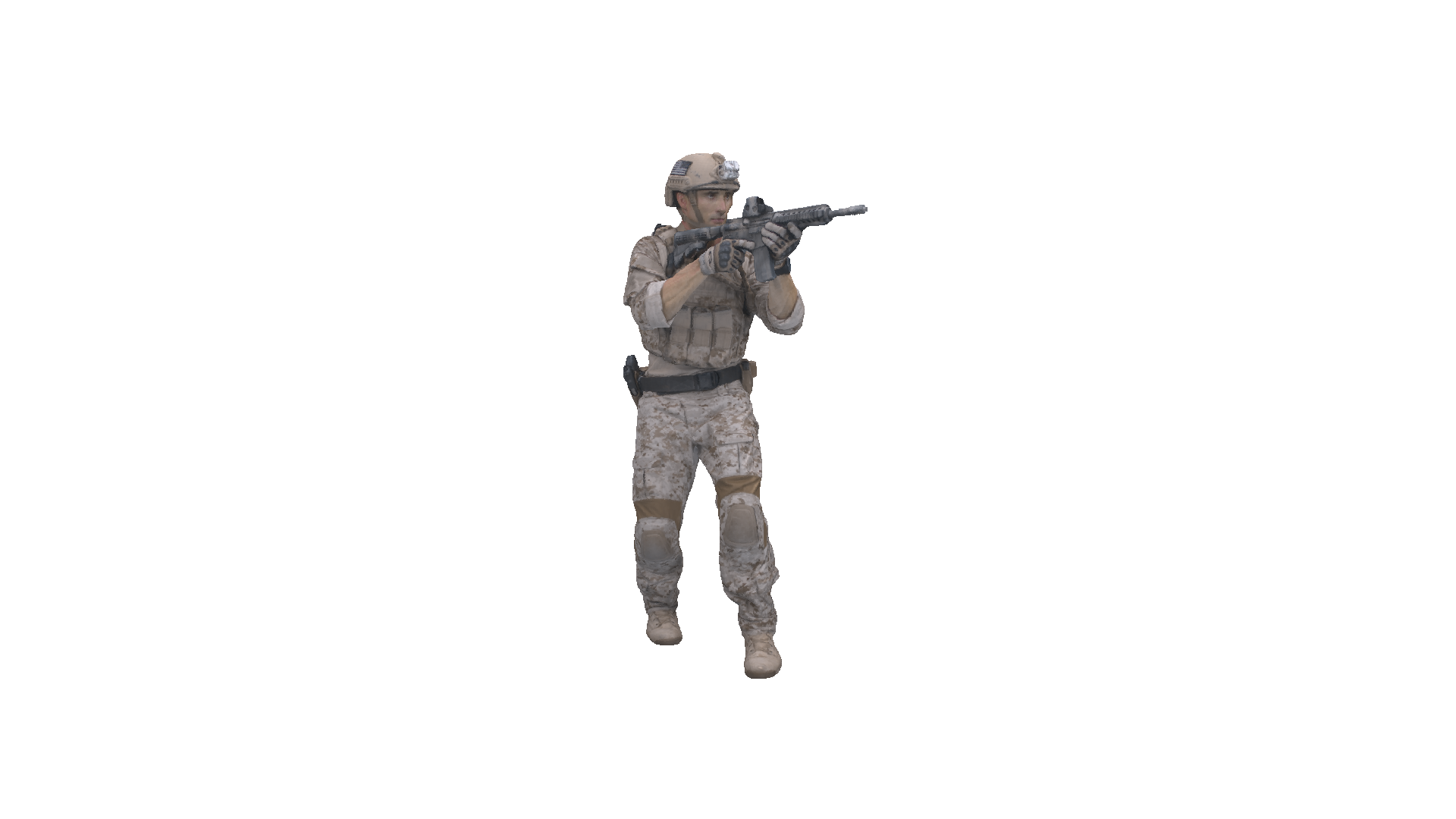} \label{fig:soldier_ref}}
    \subfloat[V-PCC ($0.60$ bpp)]{\includegraphics[trim={29cm 20cm 27.4cm 5cm},clip,width=0.25\columnwidth]{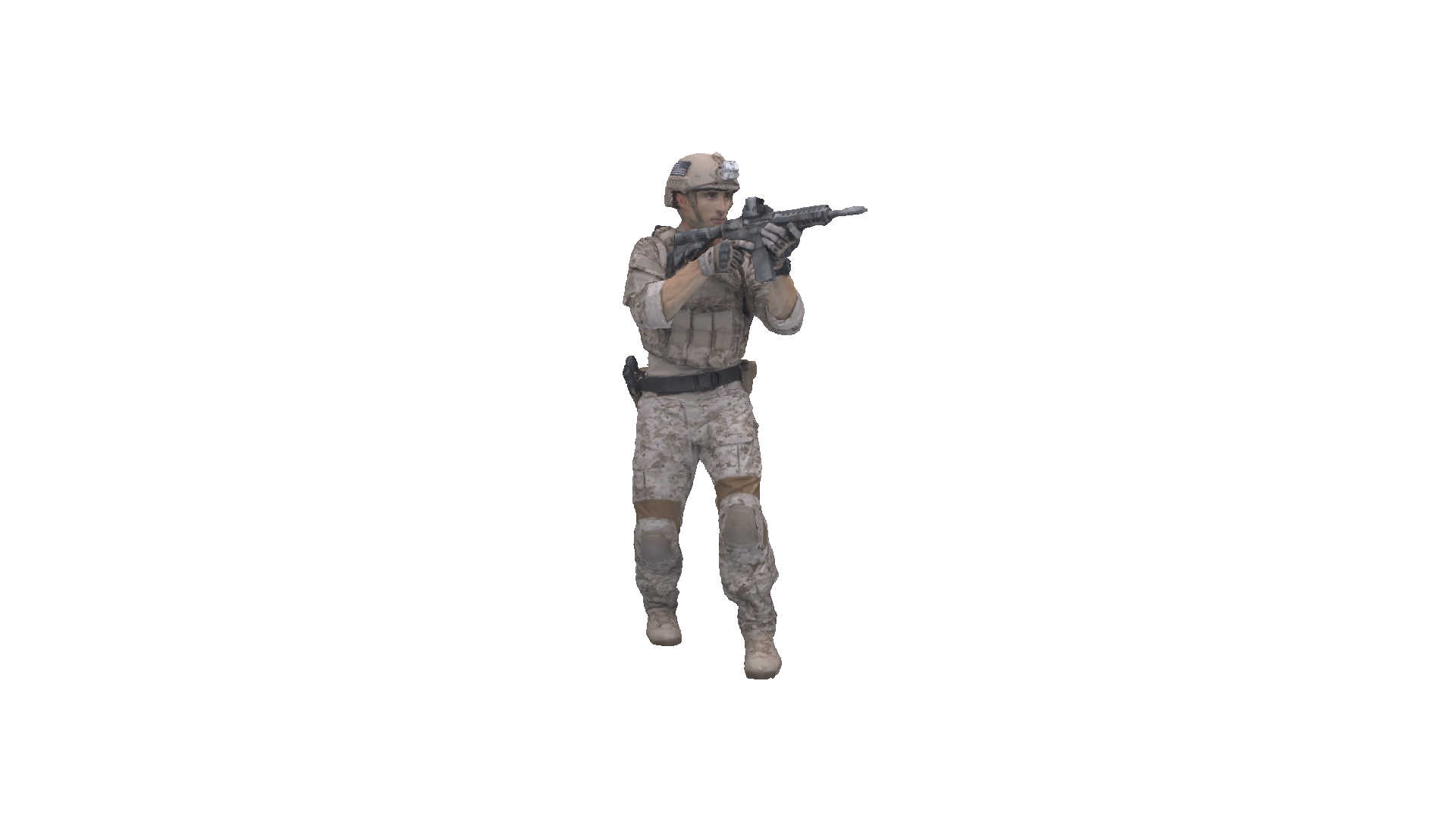} \label{fig:soldier_vpcc}}
    \subfloat[G-PCC ($0.62$ bpp)]{\includegraphics[trim={29cm 20cm 27.4cm 5cm},clip,width=0.25\columnwidth]{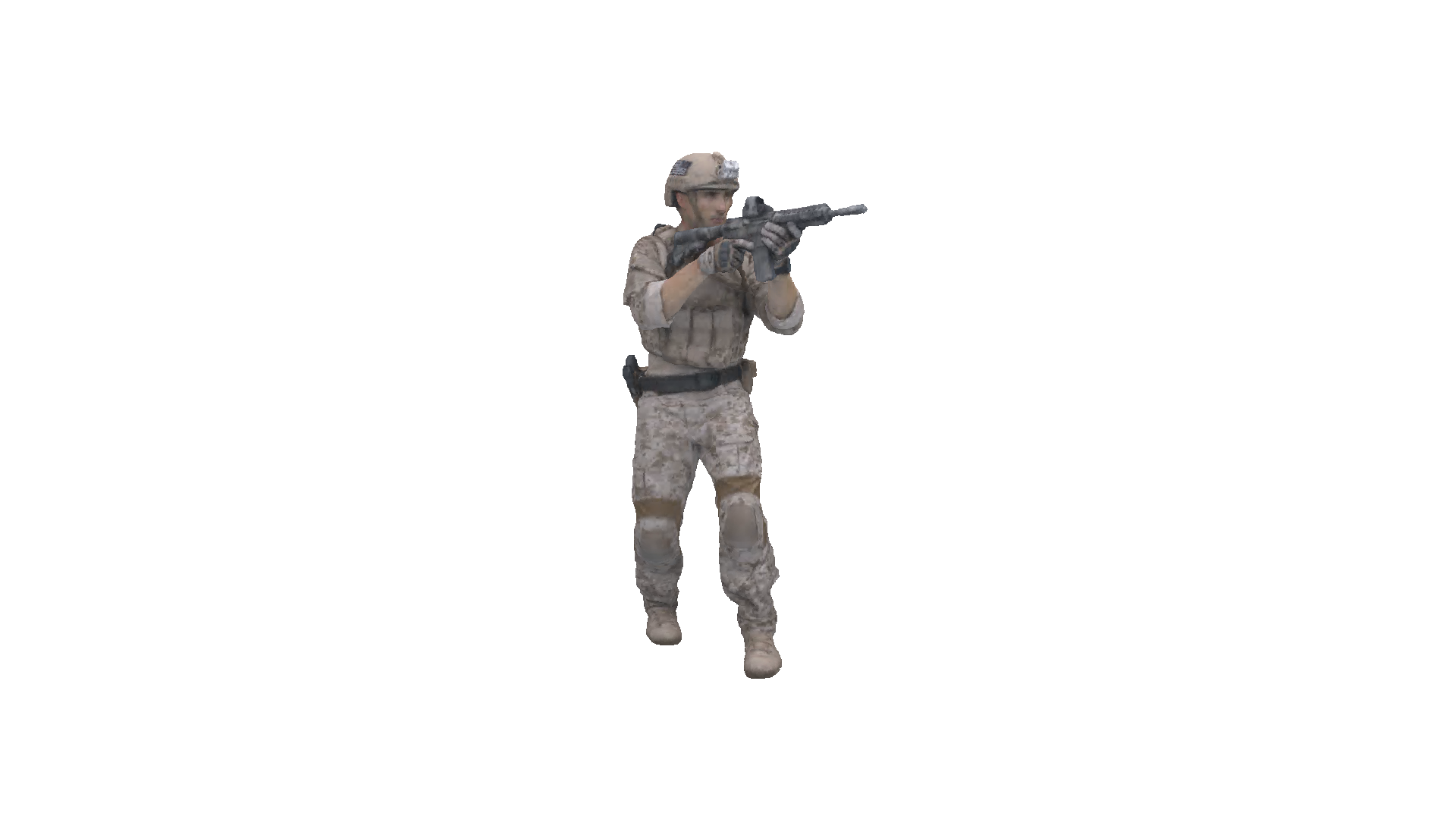} \label{fig:soldier_gpcc}}
    \subfloat[Ours ($0.61$ bpp)]{\includegraphics[trim={29cm 20cm 27.4cm 5cm},clip,width=0.25\columnwidth]{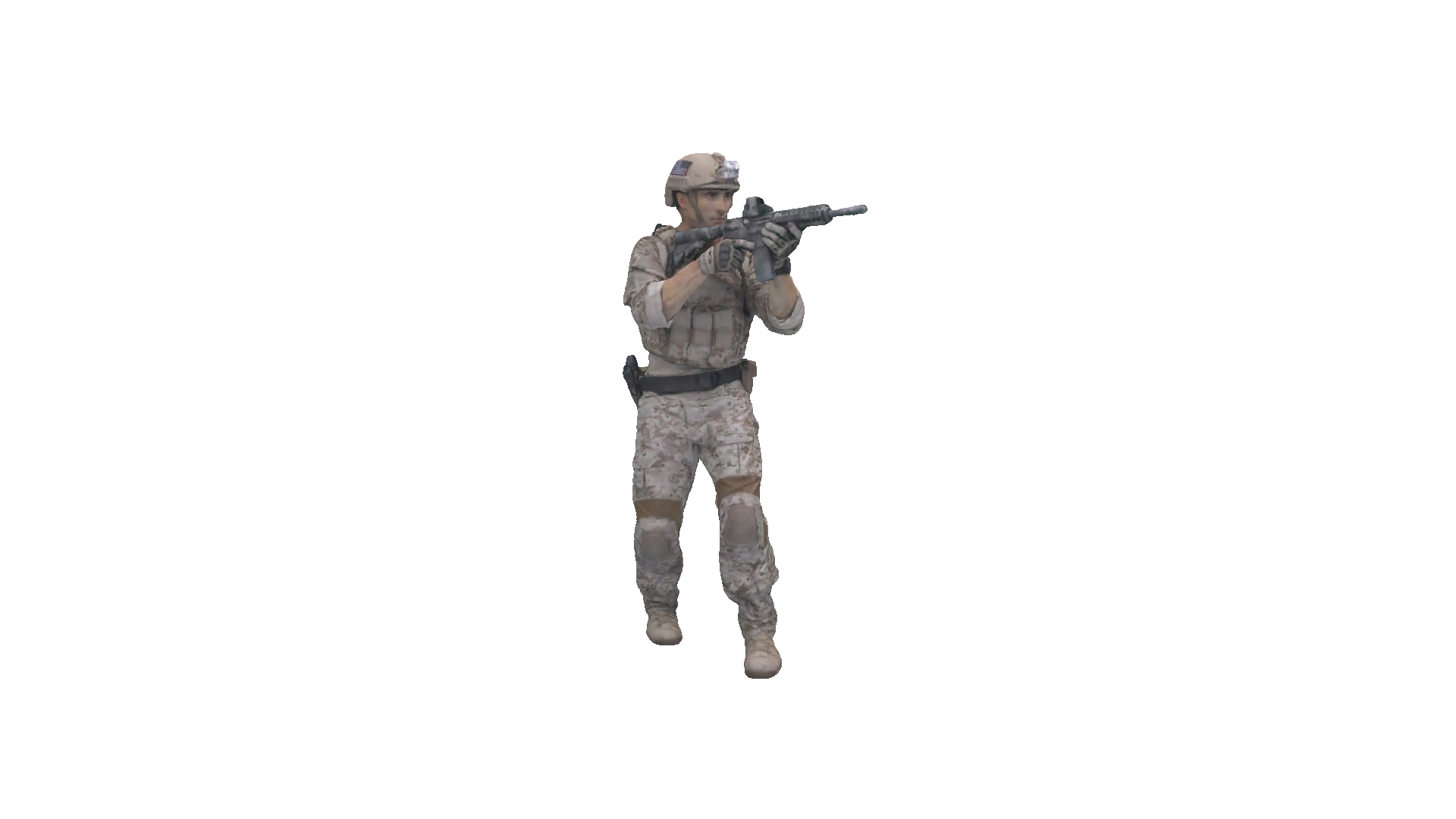} \label{fig:soldier_ours}}
    \caption{Renders of point cloud \textit{soldier}~\citep{upperBodies20178i} at the same rate.}
    \label{fig:visual results}
\end{figure}

\subsection{Latency Analysis}
\label{sec:latency}
Using a single, joint model in our approach allows to significantly simplify the encoding process, only requiring a single pass through the encoder. 
This stands in contrast to the standard procedure, which requires projection of the attributes to the reconstructed geometry during attribute encoding, as it is done in G-PCC~\citep{gpcc2021}, V-PCC~\citep{vpcc2021} and YOGA~\citep{zhang2023yoga}.
Consequently, the encoding latency of our model ($0.75s$) in Table~\ref{tab:latency} is more then one order of magnitude lower then the one for the implementations of traditional codecs~\citep{gpcc2021, vpcc2021}. 
When comparing to the learned approach in YOGA~\citep{zhang2023yoga}, we also notice a lower encoding latency by one order of magnitude compared to the results reported in the paper.
We note that the original paper utilizes different hardware, however, the encoding speed advantage of our approach, due to the single pass, is apparent.

Considering the decoding latency, we find that our method is competitive with respect to the related approaches, although all methods may require further optimization to allow application in practical use cases.
Note, however, that the latency for all approaches significantly depends on the content (i.e. number of points) and the quality configuration. 
A more detailed, per-content and per-configuration analysis is given in the appendix.

\begin{table}[]
    \centering
    \caption{Comparison of complexity of related works compressing the test point clouds of the 8i dataset~\citep{upperBodies20178i}. $\dagger$ indicates results from the original paper~\citep{zhang2023yoga} on different hardware.}
    \begin{tabular}{c|rrr}
        \toprule
         Method &Enc. time &Dec. time & Model size \\
         \midrule
         G-PCC &11.24s &1.51s &n/a \\
         V-PCC &57.29s &1.76s &n/a\\
         YOGA &$^\dagger$8.15s &$^\dagger$3.23s &$^\dagger$169.5 MB \\
         Ours &0.75s &1.84s &120.1 MB \\
         \bottomrule
    \end{tabular}
    \label{tab:latency}
\end{table}

\subsection{Showcase: View-dependent Compression}
\label{sec:view}
\begin{figure}
    \centering
    \subfloat[Uniform Quality]{\includegraphics[trim={0cm 0cm 0cm 0cm},clip,width=0.33\columnwidth]{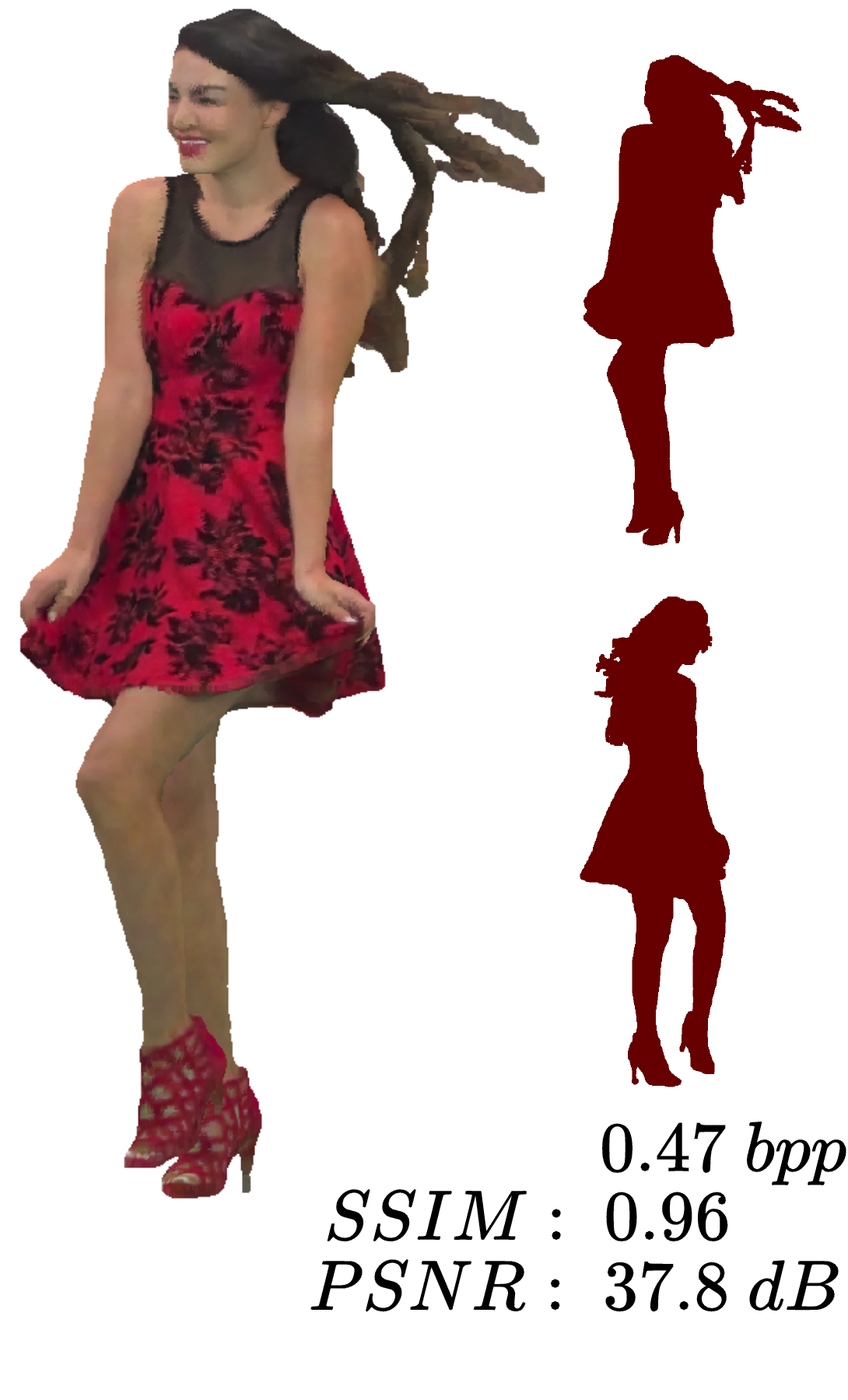} \label{fig:uni_render}}
    \subfloat[Quality Gradient]{\includegraphics[trim={0cm 0cm 0cm 0cm},clip,width=0.33\columnwidth]{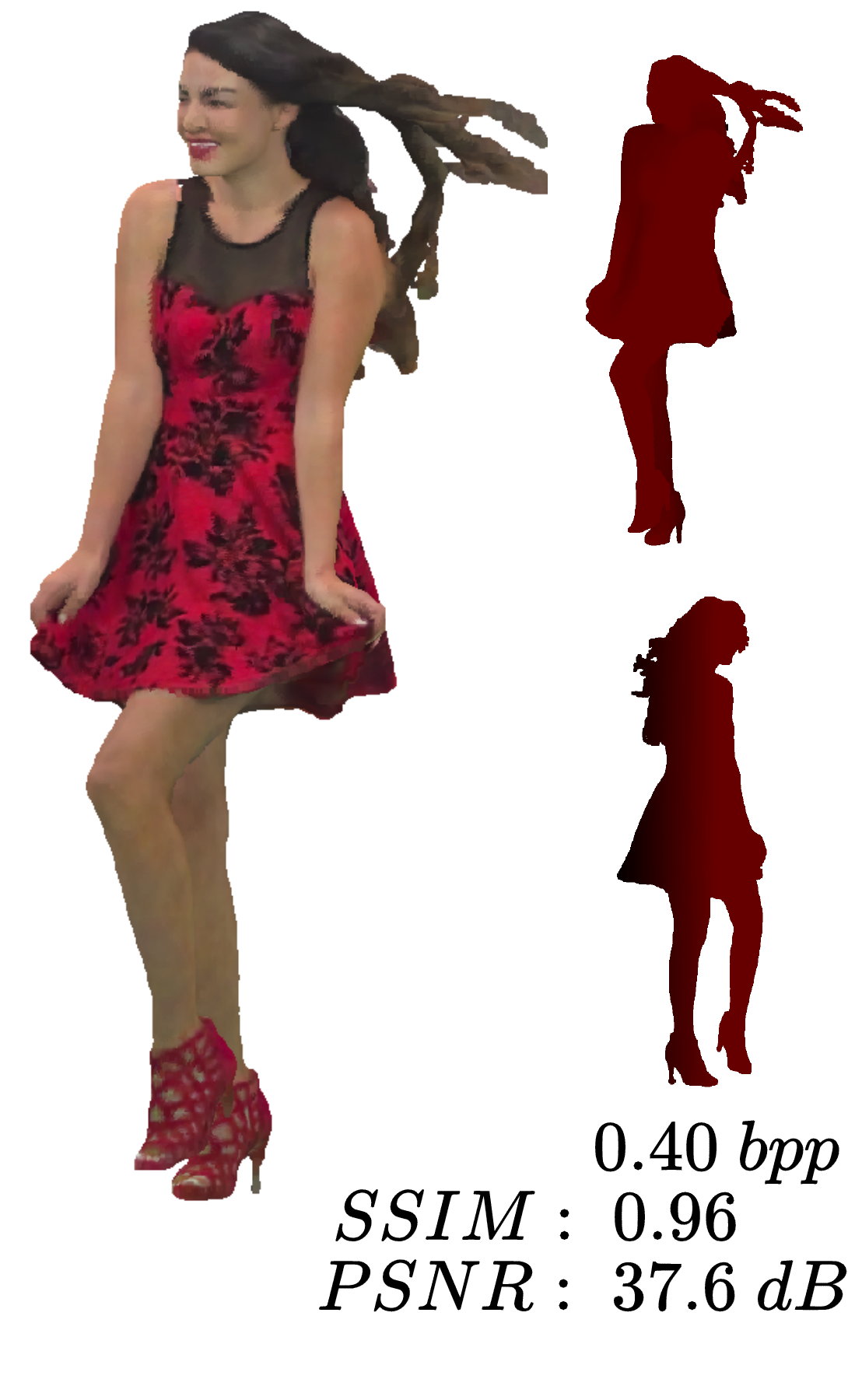} \label{fig:view_render}}
    \subfloat[Quality Cut]{\includegraphics[trim={0cm 0cm 0cm 0cm},clip,width=0.33\columnwidth]{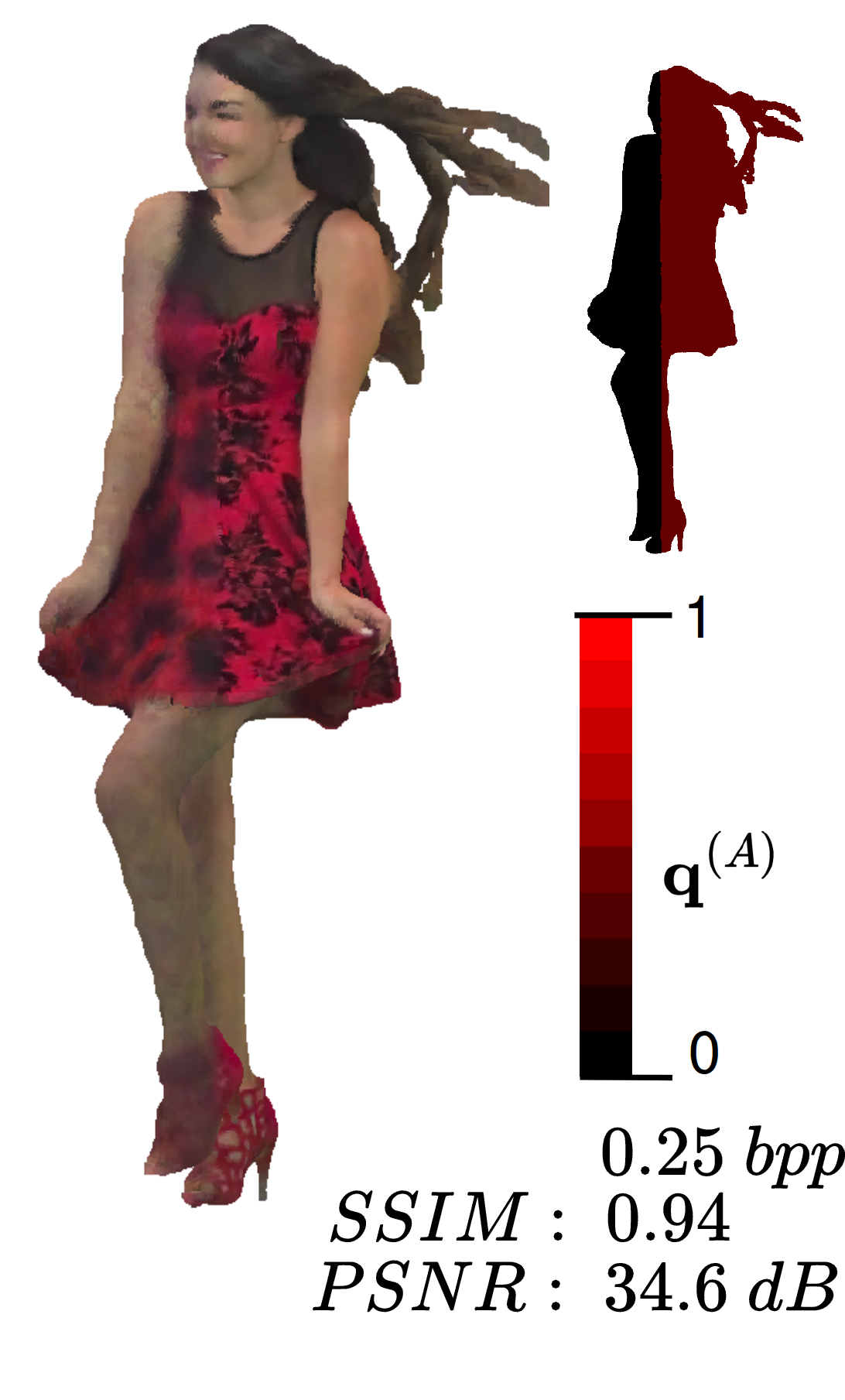} \label{fig:roi_render}}
    \caption{Rendered views of the point cloud \textit{redandblack} for different quality maps and corresponding image metrics.}
    \label{fig:viewdependent}
\end{figure}

Recall that conditioning our model on a quality map $(\mathbf{g}, \mathbf{q})$ instead of a scalar quality parameter allows to perform region-of-interest rate control.
To showcase this capability, we derive a view-dependent compression of a point cloud given some arbitrary user viewing angle.
This is achieved by instantiating quality maps using a gradient along the viewing direction, assigning higher qualities to the front of the point cloud which decreases to the back.
For comparison, we consistently apply the same procedure using a uniform quality map in Fig.~\ref{fig:uni_render} and a quality map according to the step function in Fig.~\ref{fig:roi_render}.
For all compressed point clouds in Fig.~\ref{fig:viewdependent}, the geometry quality map $\mathbf{q}^{(G)}$ is instantiated following an identical trend. 
The reconstructions are rendered from the given viewing-angle and standard image metrics, i.e. SSIM and the PSNR of the MSE in the YUV color space, are computed.
Considering the image metrics of the renders, we find some potential for reducing the rate by locally dictating the quality $\mathbf{q}$ in both image metrics, i.e., PSNR and SSIM.

\section{Conclusion and Outlook}
We presented a method for training a \textit{single, joint compression} model for point cloud geometry and attributes.
Key to our approach is extending variable-rate compression techniques to multimodal compression, allowing to gauge the model on a tradeoff between the triplet of the quality of both modalities and the rate.
Equipped with this, we proposed a joint compression architecture operating on a sparse tensor representation. Our technique achieves comparable compression performance to the state of the art in learned point cloud compression~\citep{zhang2023yoga} which, however, has significantly higher encoding latency as it requires geometry decoding at the sender to encode the attributes, while outperforming G-PCC~\citep{gpcc2021} and further closing the gap to V-PCC~\citep{vpcc2021}. 
Our adaptive, variable-rate model illustrates the local degrees of freedom and flexibility in the joint compression of geometry and attributes. 

However, compressing point clouds according to content specific configurations  allows to fully utilize our model capabilities, but requires computationally expensive grid search in the space of possible configuration pairs.
In some use cases, this may result in resorting to fixed configuration sets as evaluated.
Tackling a similar issue in video compression, per-title encoding~\citep{de2016complexity, amirpour2021pstr} offers more efficient methods for optimizing a configuration set on specific content.

Deriving meaningful quality maps to locally condition the point cloud geometry and attribute quality using our method is a promising extension of this work. 
Our evaluation gives a perspective  on the capabilities of view-dependent and region-of-interest based compression.
Possible directions are incorporating per-point importance, e.g.  through harnessing fixation maps of the content~\citep{zhou2023qava} or incorporating task-specific optimization~\citep{song2021variable}.

\bibliographystyle{unsrtnat}
\bibliography{references} 

\clearpage
\appendix

\begin{figure*}[btp]
    \centering
    \includegraphics[width=\textwidth]{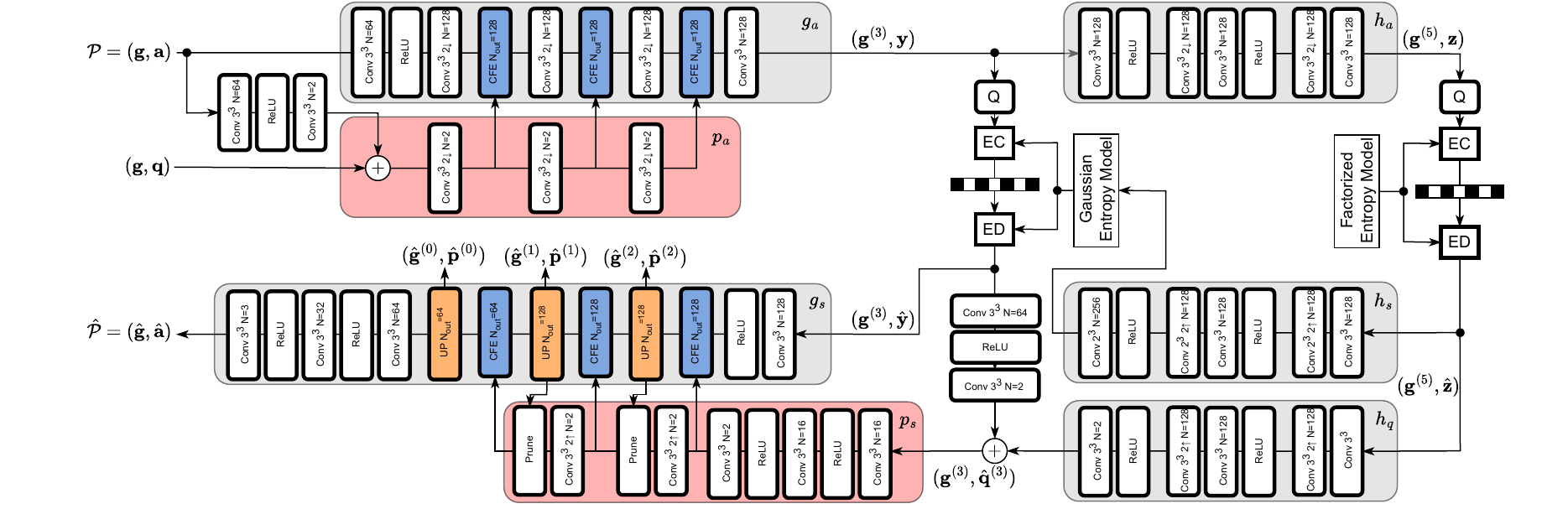}
    \caption{Implementation details for our architecture.}
    \label{fig:App_Architecture}
\end{figure*}

\section{Architecture}
We extend the visualization in Fig.~\ref{fig:architecture}, presenting a more detailed version in Fig.~\ref{fig:App_Architecture} with implementation channel sizes for reproducability. 
All convolution layers, denoted \textit{Conv} are three-dimensional sparse convolutions, for which we use Minkowski\-Engine~\citep{choy20194d}.
For layers with the extension $\uparrow$, Generative Transposed Convolutions~\citep{gwak2020gsdn} are used, allowing to generate geometry locations and corresponding features during upsampling.
Note that the hyperdecoder $h_s$ is prone to non-deterministic behaviour~\citep{balle2018integer}, which arises when using Minkowski\-Engine~\citep{gwak2020gsdn}.
To this end, we reduce the kernel size for generative convolutions in the hyperdecoder to $2$, as we found that this allows to counteract non-determinism in combination with sorting point locations before and after the convolution.

The geometry representation $\mathbf{g^{(3)}}$ at the entropy bottleneck is encoded using G-PCC in lossless geometry mode, which is common practice~\citep{wang2021lossy, wang2021multiscale, zhang2023yoga}. 
The base geometry $\mathbf{g^{(5)}}$ required to decode the hyperpriors $\mathbf{z}$ is derived through dyadic downsampling of $\mathbf{g^{(3)}}$.

In addition, the number of voxels to keep at each pruning stage in the UP block (cf.~\ref{sec:Upsampling}) is appended as side-information in the header of the bitstream.

\section{Additions to the Evaluation}
\subsection{Test Data}
The point cloud frames select for testing are presented in Tab.~\ref{tab:point_clouds}, consisting of the 8iVFBv2~\citep{upperBodies20178i} and MVUB dataset~\citep{loop2016microsoft}.
While point clouds in the 8iVFBv2 dataset~\citep{upperBodies20178i} consist of watertight surfaces and high quality geometry and textures at voxel resolution $1024$, point clouds in MVUB dataset~\citep{loop2016microsoft} are frontal scans of human upper bodies containing visible (geometry) noise at voxel resolution $512$. 
Incorporating both datasets in the evaluation allowed us to assess the performance of our model over broad input conditions.

\begin{table}[]
\centering
\caption{Tested point clouds}
\begin{tabular}{llll}
\toprule
Dataset                  & Sequence                              & Frame & Resolution \\
\midrule
\multirow{4}{*}{8iVFBv2~\citep{upperBodies20178i}} & \textit{longdress}   & 1300  & 1024       \\
                         & \textit{soldier}     & 690   & 1024       \\
                         & \textit{loot}        & 1200  & 1024       \\
                         & \textit{redandblack} & 1550  & 1024       \\
\midrule
\multirow{4}{*}{MVUB~\citep{loop2016microsoft}}    & \textit{phil9}       & 0     & 512        \\
                         & \textit{sarah9}      & 0     & 512        \\
                         & \textit{david9}      & 0     & 512        \\
                         & \textit{andrew9}     & 0     & 512        \\
\bottomrule
\end{tabular}
\label{tab:point_clouds}
\end{table}

\subsection{Extended Results for the Latency}
Complementary to the results presented in Table~\ref{tab:latency} in Sec.~\ref{sec:latency}, we showcase per content results of the encoding latency in Table~\ref{tab:enc_latency} and decoding latency in Table~\ref{tab:dec_latency}.
Since YOGA~\cite{zhang2023yoga} is not open-sourced at the time of our evaluation, we can only report the averaged latencies from their results, which were measured on different hardware.
For all other approaches, we use the configurations for the generation of the pareto-fronts in Sec.~\ref{sec:pareto} and measure the times for each encoding and decoding.
We notice that the encoding and decoding latency of G-PCC~\cite{gpcc2021} heavily relies on the quality configuration for the geometry and thus split the results for each tested geometry setting. 
It becomes apparent that the latency highly depends on the content, more specifically on the number of points. 

Overall, we observe similar behaviour as presented in Sec.~\ref{sec:latency}: Our method allows significantly faster encoding times as other methods, which can be mainly attributed to the fact that the geometry does not need to be reconstructed for attribute re-projection during the encoding process. 
For decoding, we find our method performing comparable to related work.

 \begin{table*}[]
     \centering
     \begin{tabular}{l|rrrr|r}
     \toprule
      Method   & \multicolumn{5}{c}{Encoding Latency [s]} \\
              & Loot & Redandblack & Soldier & Longdress & Average \\
     \midrule
     G-PCC r1 &$6.90\pm0.005$      &$6.56\pm0.007$           &$9.49\pm0.008$         &$7.30\pm0.014$           &$7.57$  \\
     G-PCC r2 &$5.85\pm0.006$      &$5.52\pm0.008$           &$8.03\pm0.008$         &$6.21\pm0.011$           &$6.41$  \\
     G-PCC r3 &$8.50\pm0.008$      &$8.03\pm0.011$           &$11.64\pm0.011$        &$7.34\pm0.009$           &$9.30$  \\
     G-PCC r4 &$13.64\pm0.014$     &$12.87\pm0.014$          &$18.60\pm0.024$        &$9.03\pm0.019$           &$14.90$  \\
     V-PCC    &$55.08\pm0.513$     &$56.14\pm0.498$          &$65.60\pm0.732$        &$52.3\pm0.772$           &$57.28$ \\
     YOGA     &n/a      &n/a             &n/a         &n/a           &$8.15$        \\
     Ours     &$0.66\pm0.006$      &$0.66\pm0.008$             &$0.95\pm0.011$         &$0.74\pm0.007$           &$0.75$     \\
     \bottomrule
     \end{tabular}
     \caption{Encoding latency measured on a AMD EPYC 7542 using a NVIDIA RTX4090 graphics card over all configurations on the point clouds of the 8i dataset~\citep{upperBodies20178i} listed in the paper. $\dagger$ indicates that the results are obtained from the original paper~\citep{zhang2023yoga}, as the model is not open-sourced.}
     \label{tab:enc_latency}
\end{table*}

 \begin{table*}[]
     \centering
     \begin{tabular}{l|rrrr|r}
     \toprule
      Method   & \multicolumn{5}{c}{Decoding Latency [s]} \\
              & Loot & Redandblack & Soldier & Longdress & Average \\
     \midrule
     G-PCC r1 &$0.08\pm0.002$      &$0.07\pm0.002$             &$0.10\pm0.001$         &$0.08\pm0.001$           &$0.08$                                     \\
     G-PCC r2 &$0.28\pm0.002$      &$0.27\pm0.002$             &$0.38\pm0.002$         &$0.30\pm0.003$           &$0.31$                                     \\
     G-PCC r3 &$1.06\pm0.004$      &$1.01\pm0.004$             &$1.45\pm0.005$         &$1.14\pm0.005$           &$1.16$                                     \\
     G-PCC r4 &$2.29\pm0.005$      &$2.18\pm0.007$             &$3.14\pm0.009$         &$2.46\pm0.008$           &$2.52$                                     \\
     V-PCC     &$1.60\pm0.021$      &$1.59\pm0.018$             &$2.12\pm0.023$         &$1.71\pm0.022$           &$1.76$                                  \\
     YOGA     &n/a    &n/a           &n/a           &n/a            &$3.23$                                   \\ 
     Ours       &$1.66\pm0.009$      &$1.71\pm0.009$             &$2.23\pm0.009$         &$1.81\pm0.009$           &$1.84$ \\
     \bottomrule
     \end{tabular}
     \caption{Comparison of decoding latency, measured on a AMD EPYC 7542 using a NVIDIA RTX4090 graphics card over all configurations on the point clouds of the 8i dataset~\citep{upperBodies20178i} listed in the paper. $\dagger$ indicates that the results are obtained from the original paper~\citep{zhang2023yoga}, as the model is not open-sourced.}
     \label{tab:dec_latency}
\end{table*} 

\subsection{Complementary Rate Distortion Results for Fixed-Configuration Encoding}
\begin{figure*}
    \centering
    \subfloat{\includegraphics[trim={0cm 0cm 0cm 0cm},clip,width=0.245\textwidth]{figures/RD-Figures/rd-config_sym_p2p_psnr_loot.pdf}}
    \subfloat{\includegraphics[trim={0cm 0cm 0cm 0cm},clip,width=0.245\textwidth]{figures/RD-Figures/rd-config_sym_p2p_psnr_redandblack.pdf}}
    \subfloat{\includegraphics[trim={0cm 0cm 0cm 0cm},clip,width=0.245\textwidth]{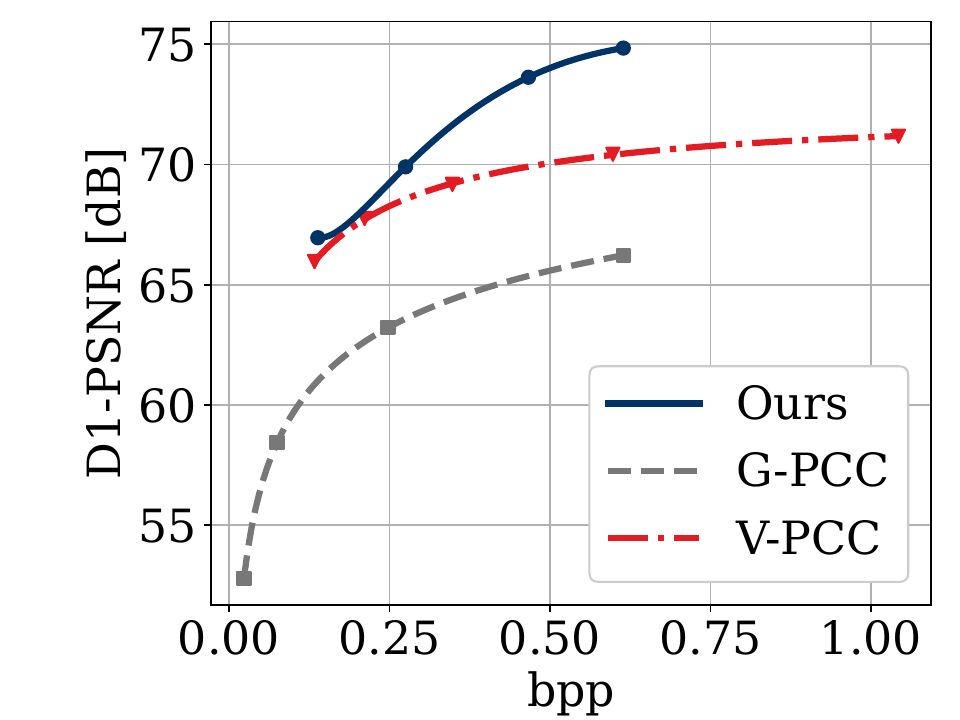}}
    \subfloat{\includegraphics[trim={0cm 0cm 0cm 0cm},clip,width=0.245\textwidth]{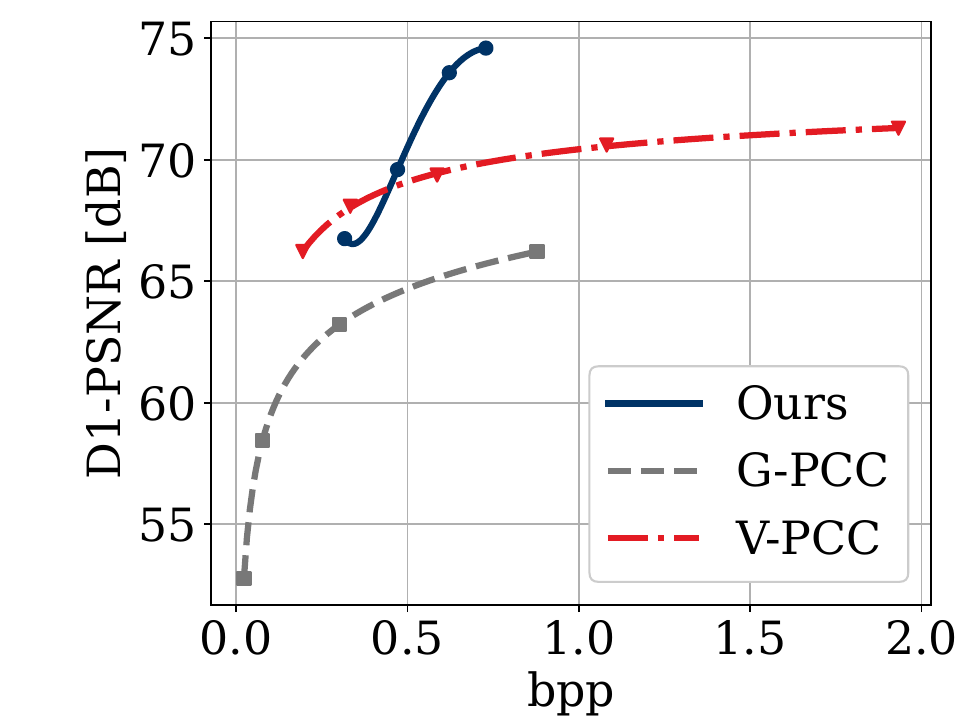}}

    \subfloat{\includegraphics[trim={0cm 0cm 0cm 0cm},clip,width=0.245\textwidth]{figures/RD-Figures/rd-config_sym_y_psnr_loot.pdf}}
    \subfloat{\includegraphics[trim={0cm 0cm 0cm 0cm},clip,width=0.245\textwidth]{figures/RD-Figures/rd-config_sym_y_psnr_redandblack.pdf}}
    \subfloat{\includegraphics[trim={0cm 0cm 0cm 0cm},clip,width=0.245\textwidth]{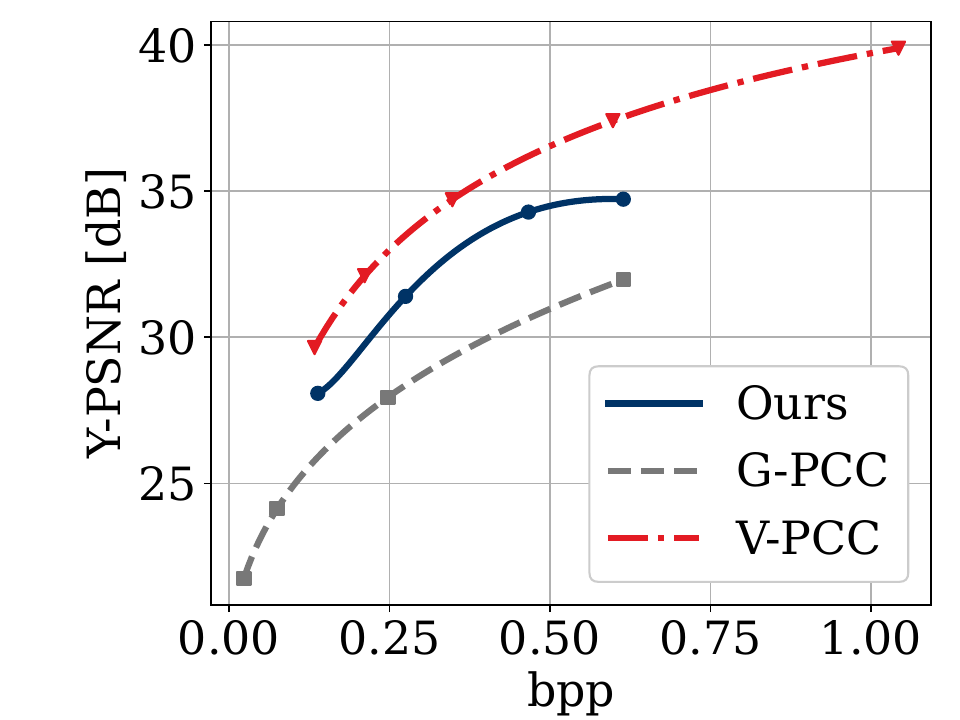}}
    \subfloat{\includegraphics[trim={0cm 0cm 0cm 0cm},clip,width=0.245\textwidth]{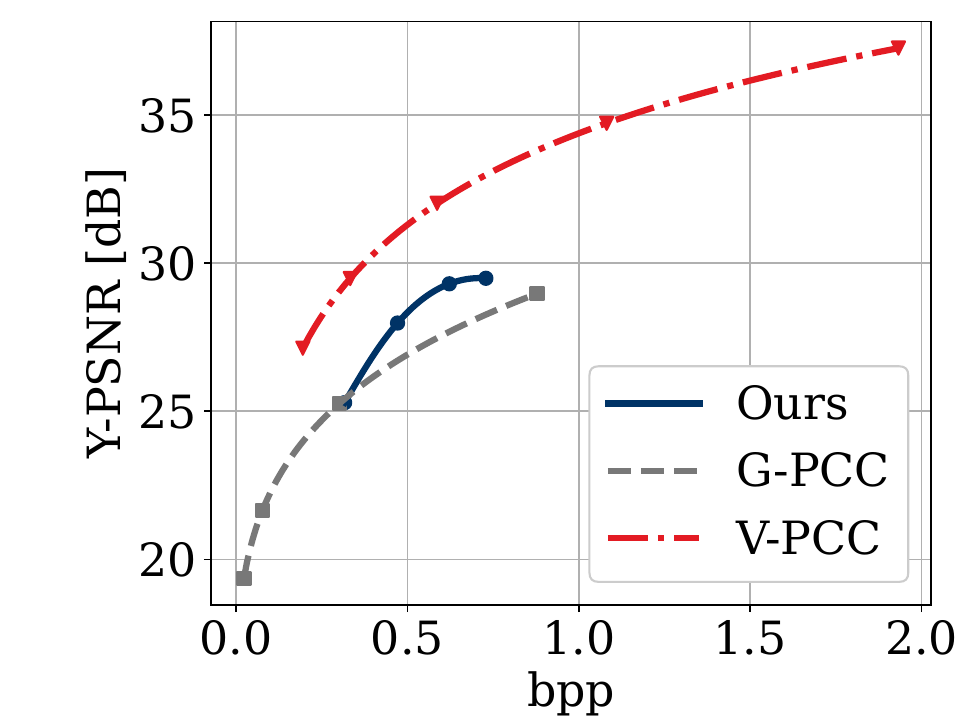}}
    
    \subfloat{\includegraphics[trim={0cm 0cm 0cm 0cm},clip,width=0.245\textwidth]{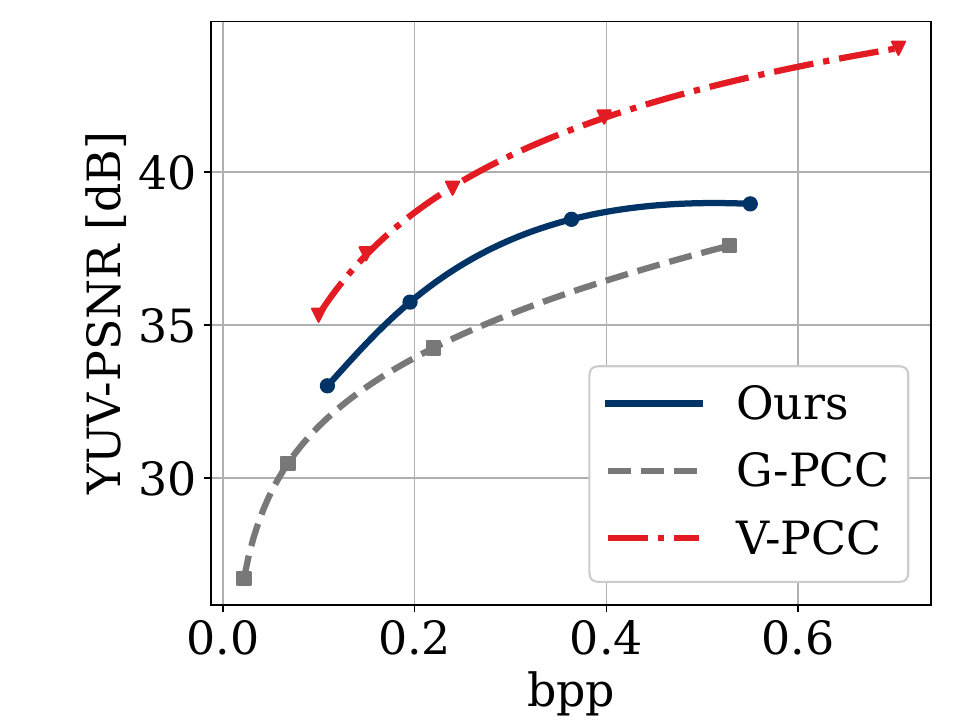}}
    \subfloat{\includegraphics[trim={0cm 0cm 0cm 0cm},clip,width=0.245\textwidth]{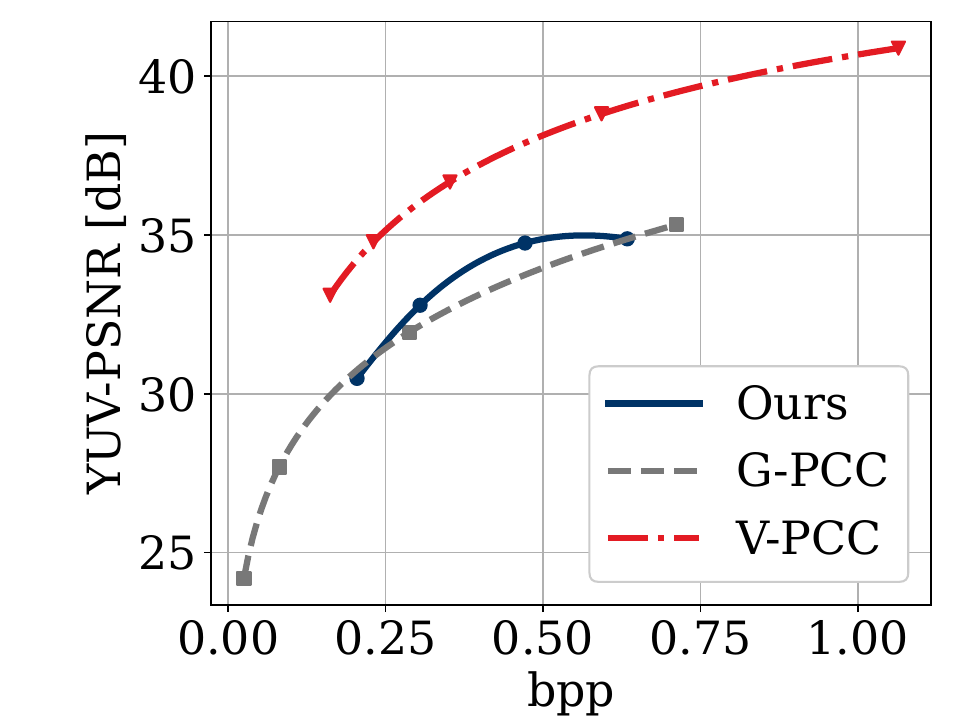}}
    \subfloat{\includegraphics[trim={0cm 0cm 0cm 0cm},clip,width=0.245\textwidth]{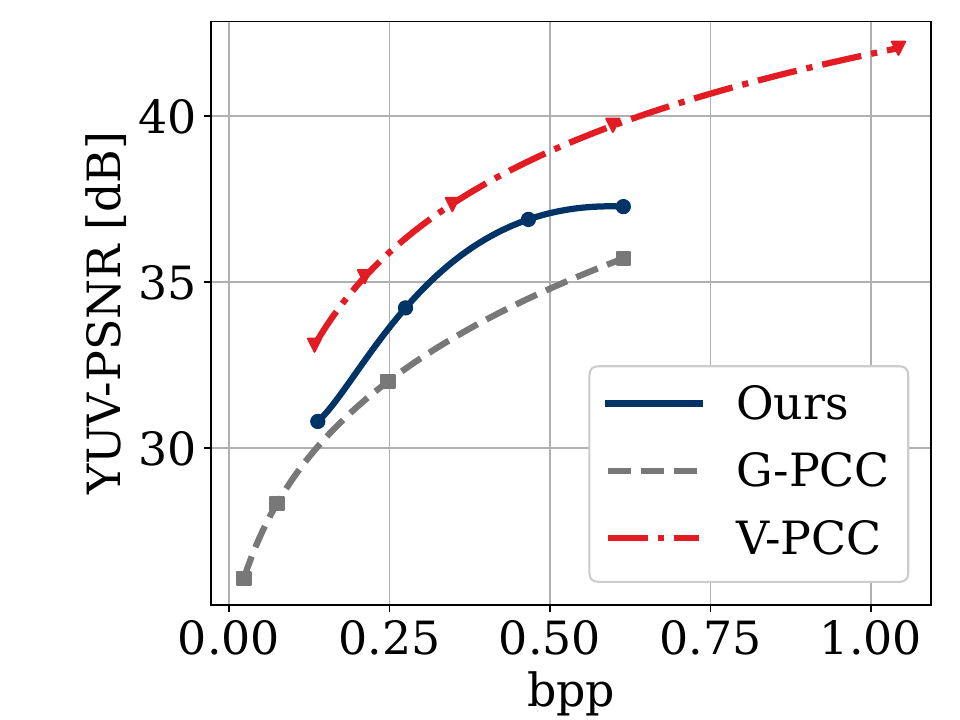}}
    \subfloat{\includegraphics[trim={0cm 0cm 0cm 0cm},clip,width=0.245\textwidth]{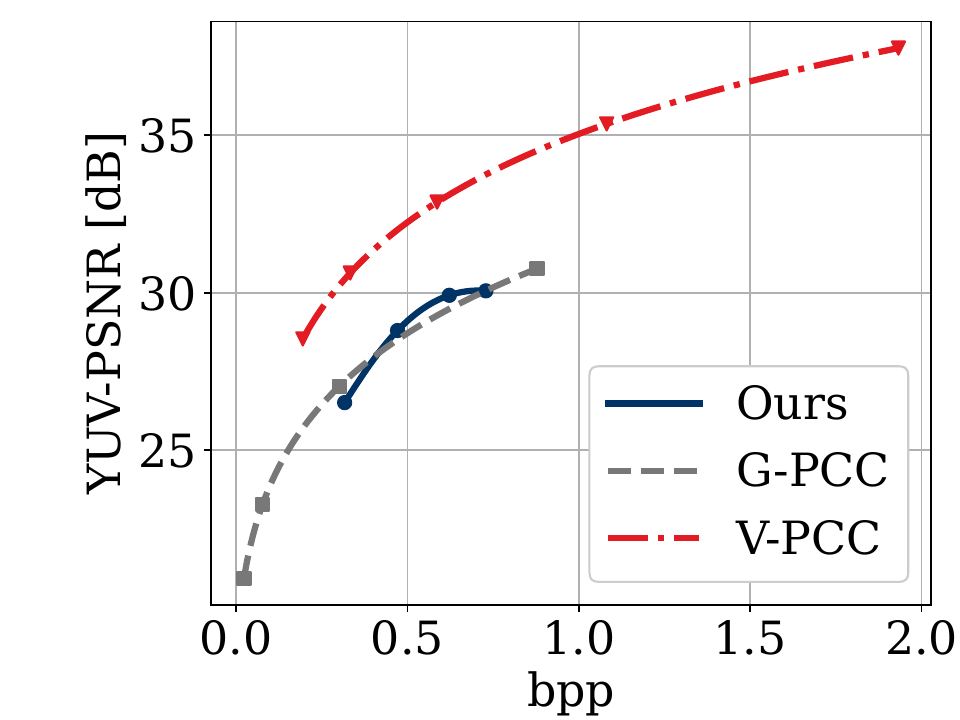}}
    \\
    
    \setcounter{subfigure}{0}
    \subfloat[\textit{loot}]{\includegraphics[trim={0cm 0cm 0cm 0cm},clip,width=0.245\textwidth]{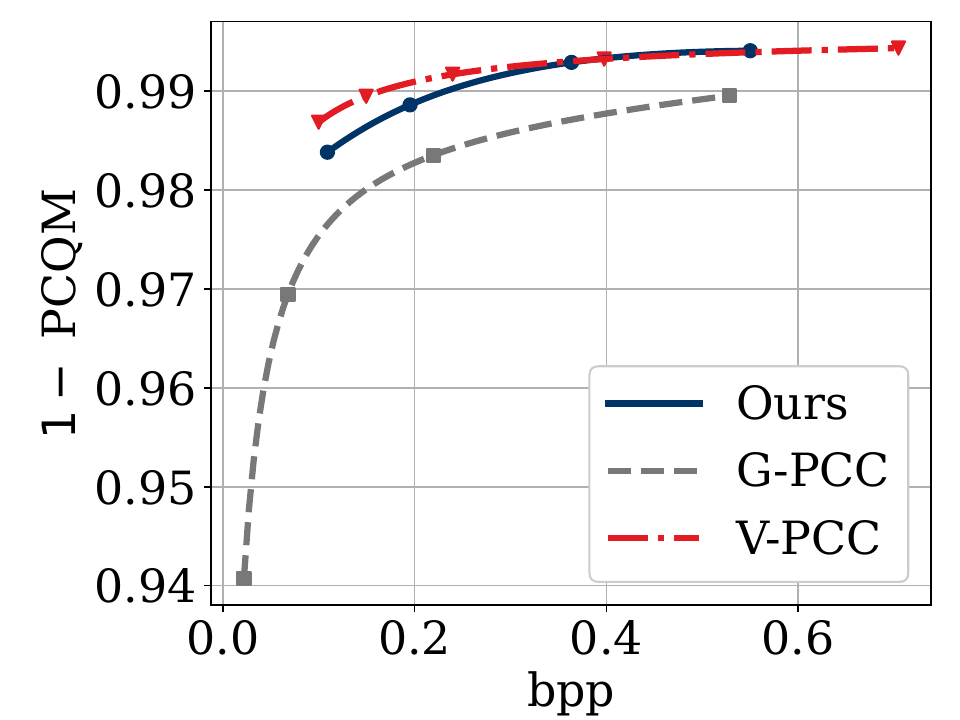} }
    \subfloat[\textit{redandblack}]{\includegraphics[trim={0cm 0cm 0cm 0cm},clip,width=0.245\textwidth]{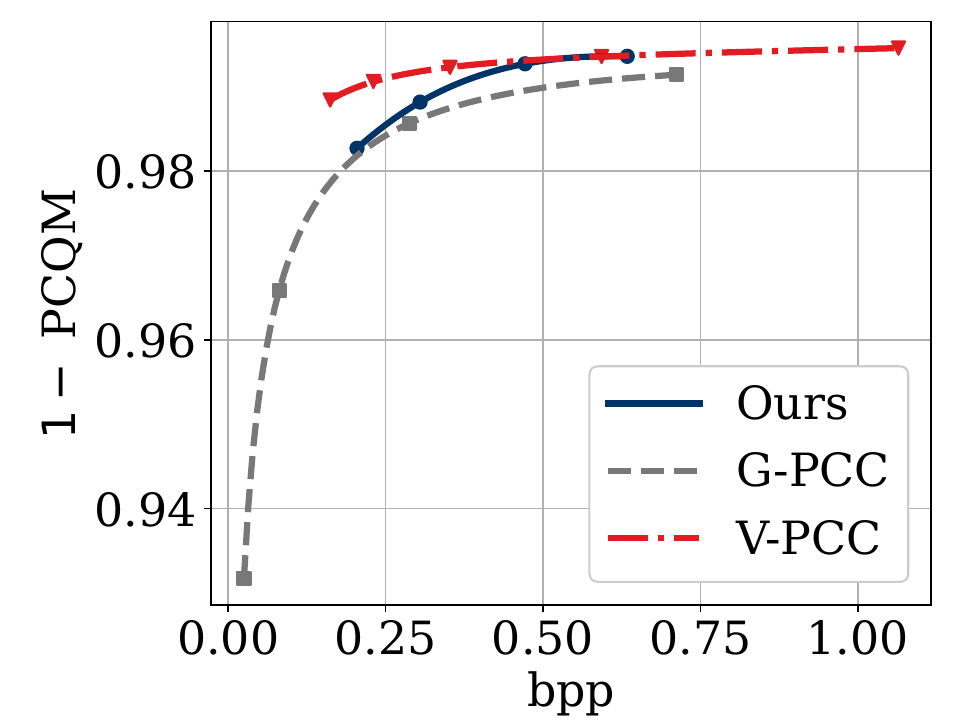}  }
    \subfloat[\textit{soldier}]{\includegraphics[trim={0cm 0cm 0cm 0cm},clip,width=0.245\textwidth]{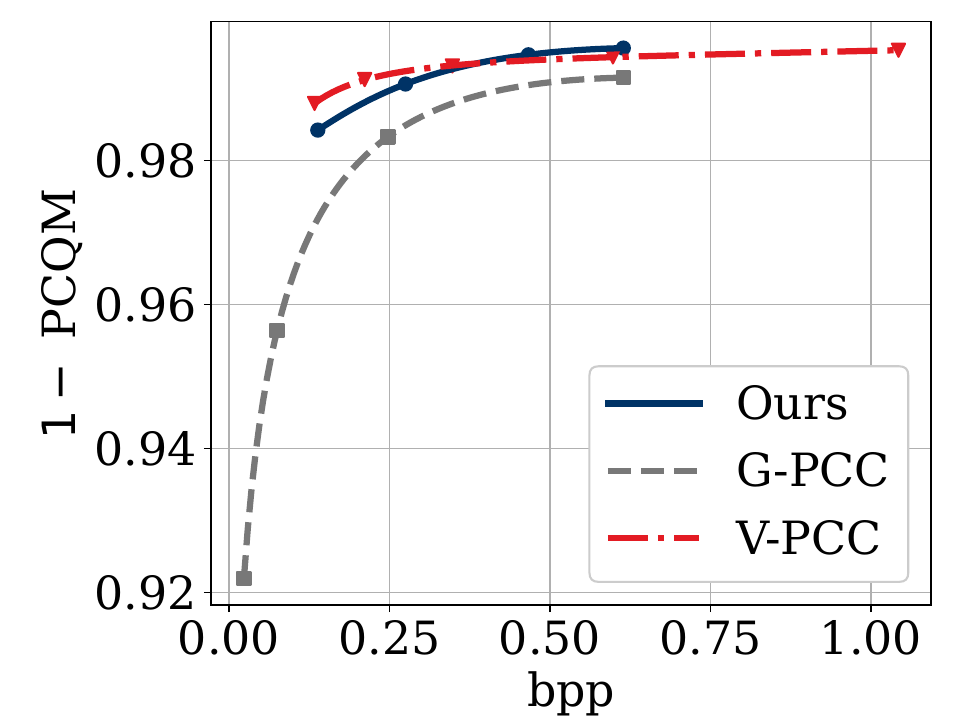} }
    \subfloat[\textit{longdress}]{\includegraphics[trim={0cm 0cm 0cm 0cm},clip,width=0.245\textwidth]{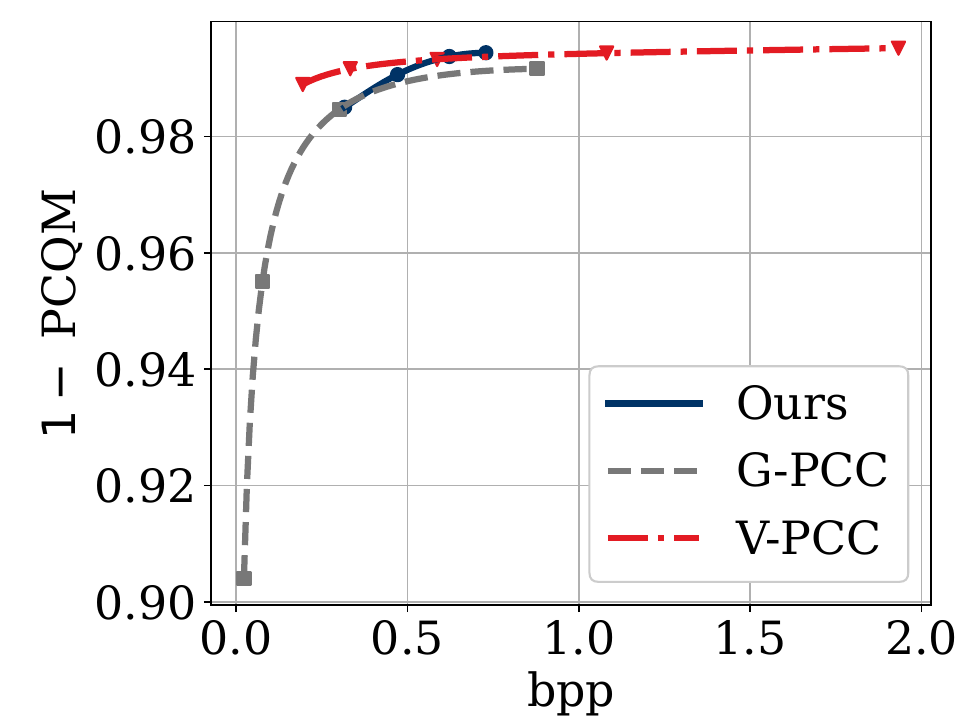} }
    \caption{Rate distortion results on the 8iVFBv2~\citep{upperBodies20178i} dataset.}
    \label{fig:appendix_8ird_results}
\end{figure*}

\begin{figure*}
    \centering
    \subfloat{\includegraphics[trim={0cm 0cm 0cm 0cm},clip,width=0.245\textwidth]{figures/RD-Figures/rd-config_sym_p2p_psnr_phil9.pdf}}
    \subfloat{\includegraphics[trim={0cm 0cm 0cm 0cm},clip,width=0.245\textwidth]{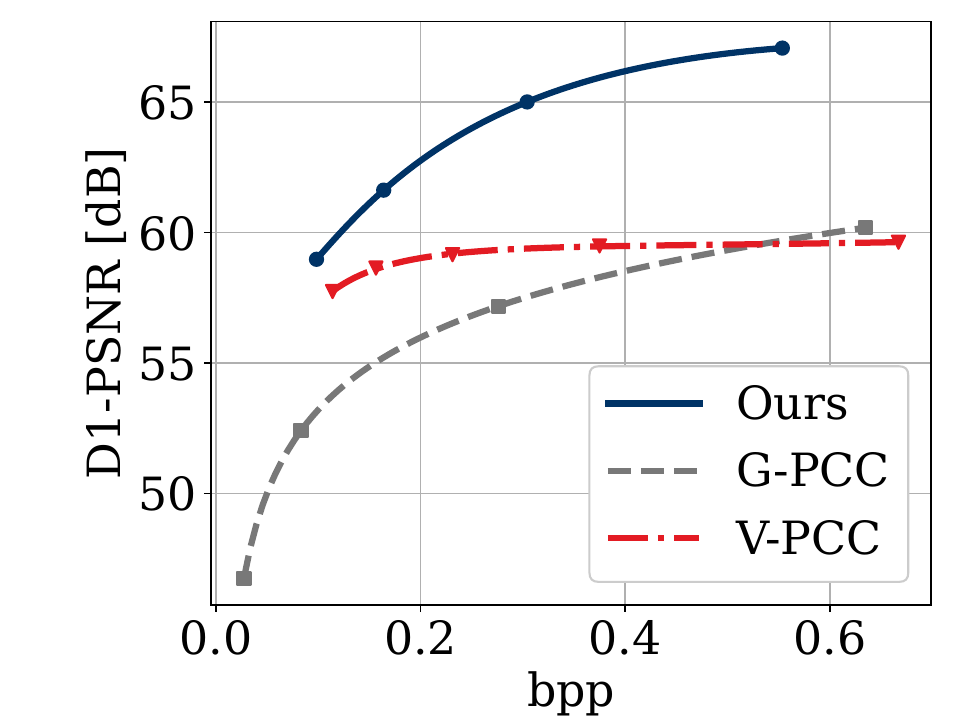}}
    \subfloat{\includegraphics[trim={0cm 0cm 0cm 0cm},clip,width=0.245\textwidth]{figures/RD-Figures/rd-config_sym_p2p_psnr_andrew9.pdf}}
    \subfloat{\includegraphics[trim={0cm 0cm 0cm 0cm},clip,width=0.245\textwidth]{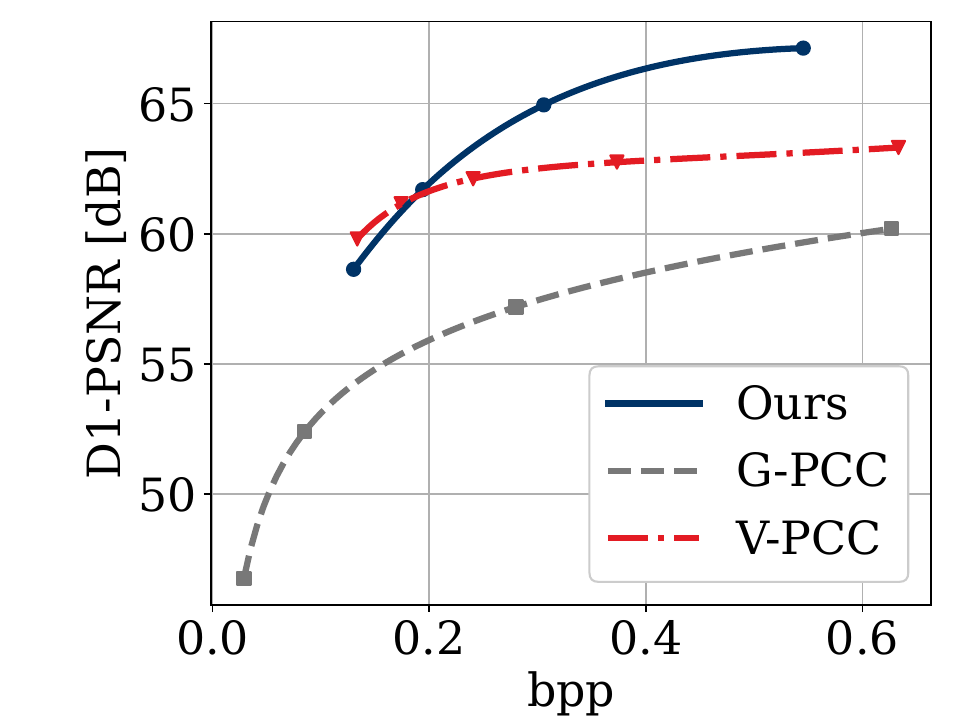}}

    \subfloat{\includegraphics[trim={0cm 0cm 0cm 0cm},clip,width=0.245\textwidth]{figures/RD-Figures/rd-config_sym_y_psnr_phil9.pdf}}
    \subfloat{\includegraphics[trim={0cm 0cm 0cm 0cm},clip,width=0.245\textwidth]{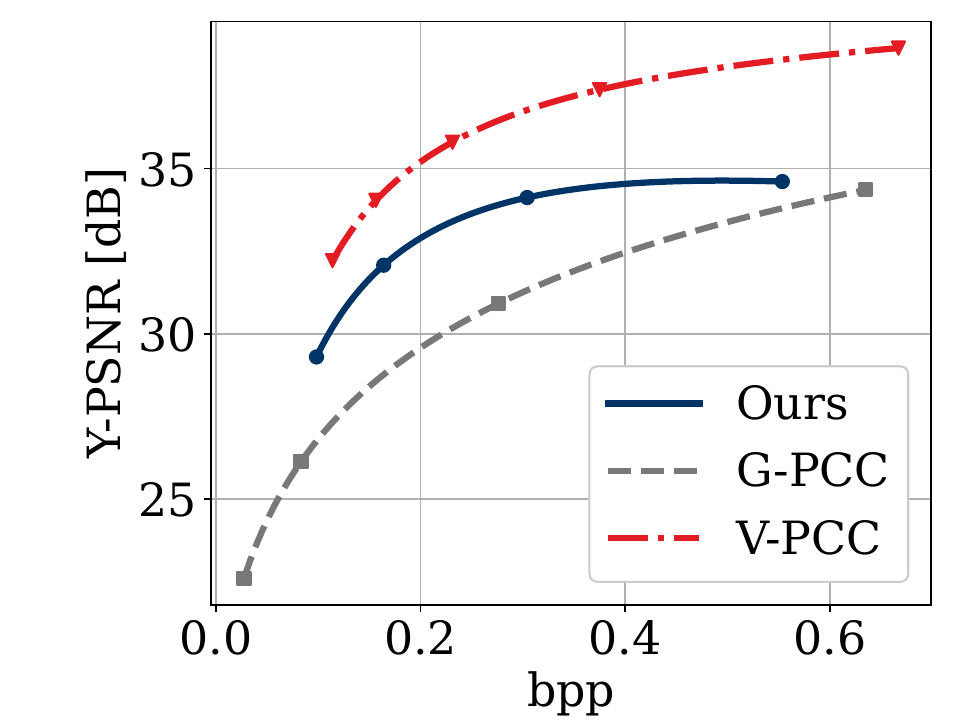}}
    \subfloat{\includegraphics[trim={0cm 0cm 0cm 0cm},clip,width=0.245\textwidth]{figures/RD-Figures/rd-config_sym_y_psnr_andrew9.pdf}}
    \subfloat{\includegraphics[trim={0cm 0cm 0cm 0cm},clip,width=0.245\textwidth]{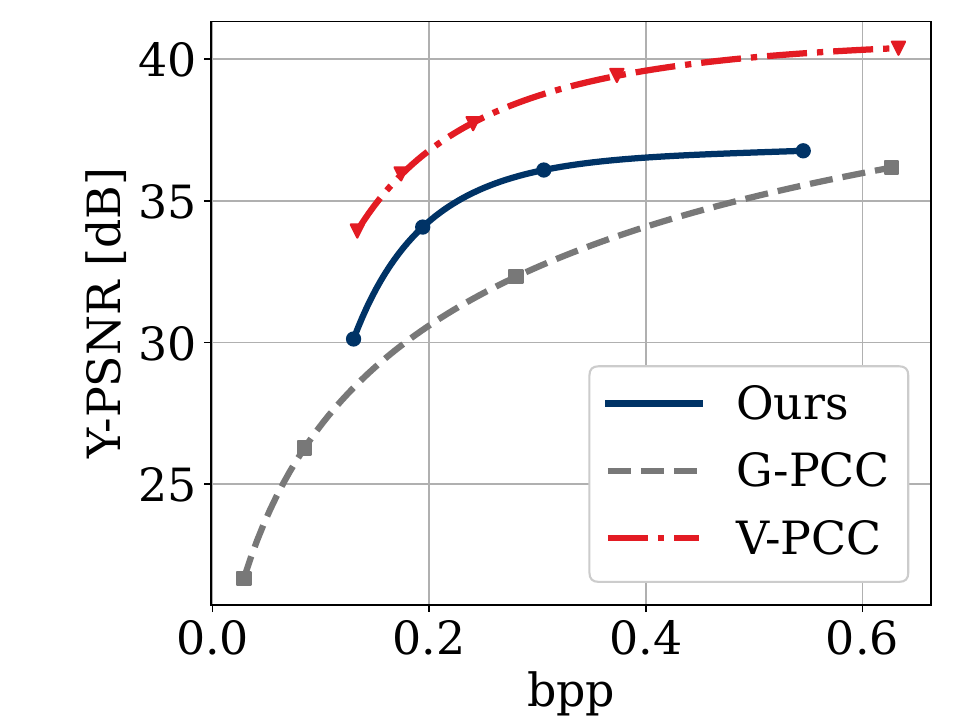}}
    
    \subfloat{\includegraphics[trim={0cm 0cm 0cm 0cm},clip,width=0.245\textwidth]{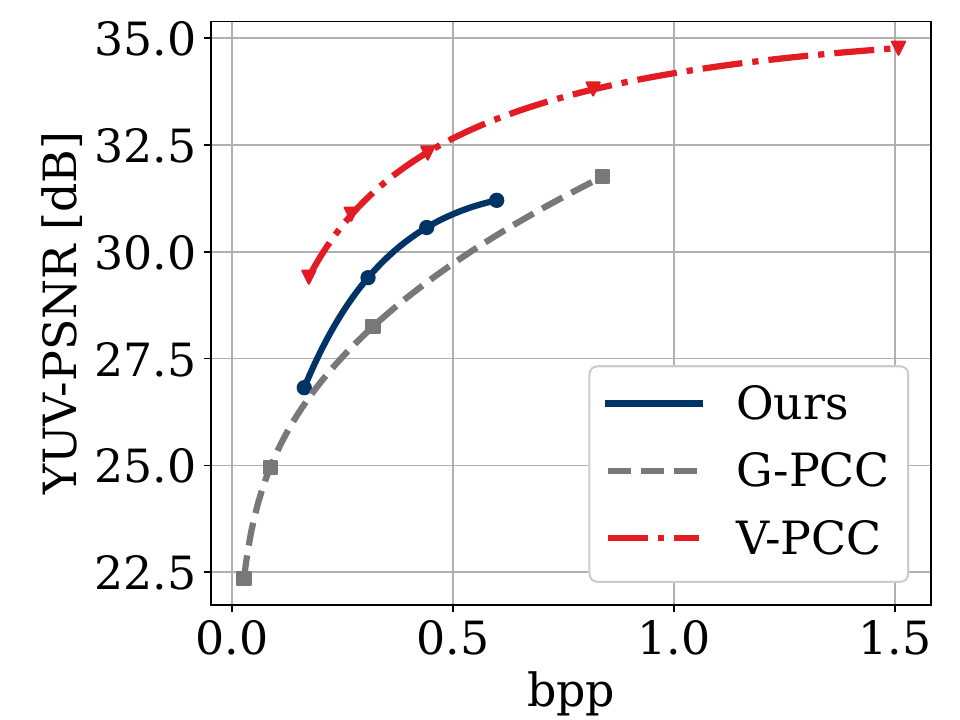}}
    \subfloat{\includegraphics[trim={0cm 0cm 0cm 0cm},clip,width=0.245\textwidth]{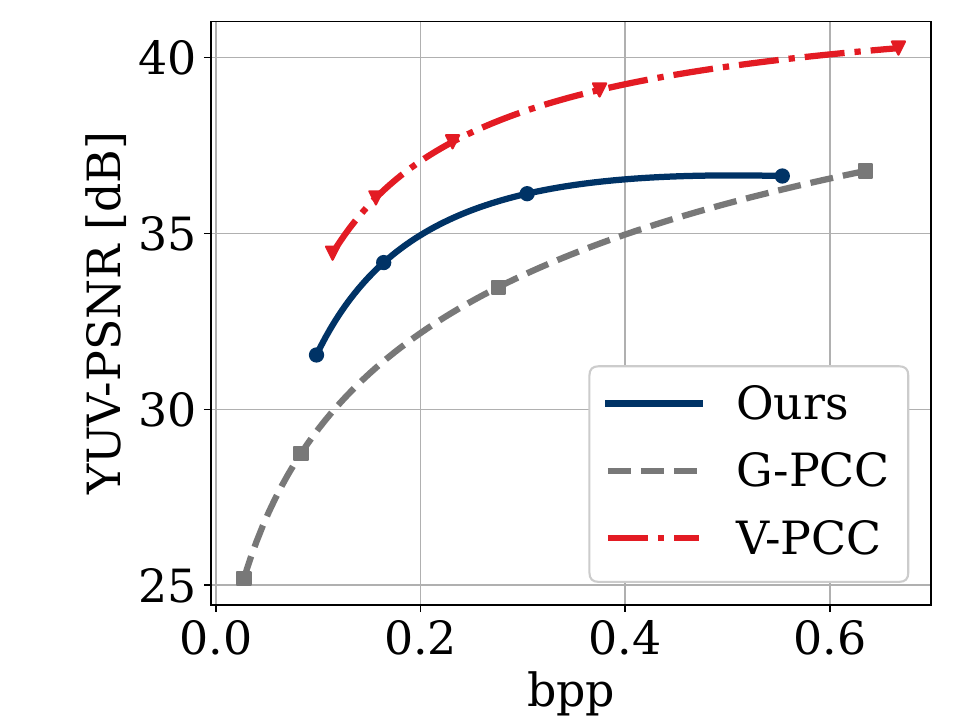}}
    \subfloat{\includegraphics[trim={0cm 0cm 0cm 0cm},clip,width=0.245\textwidth]{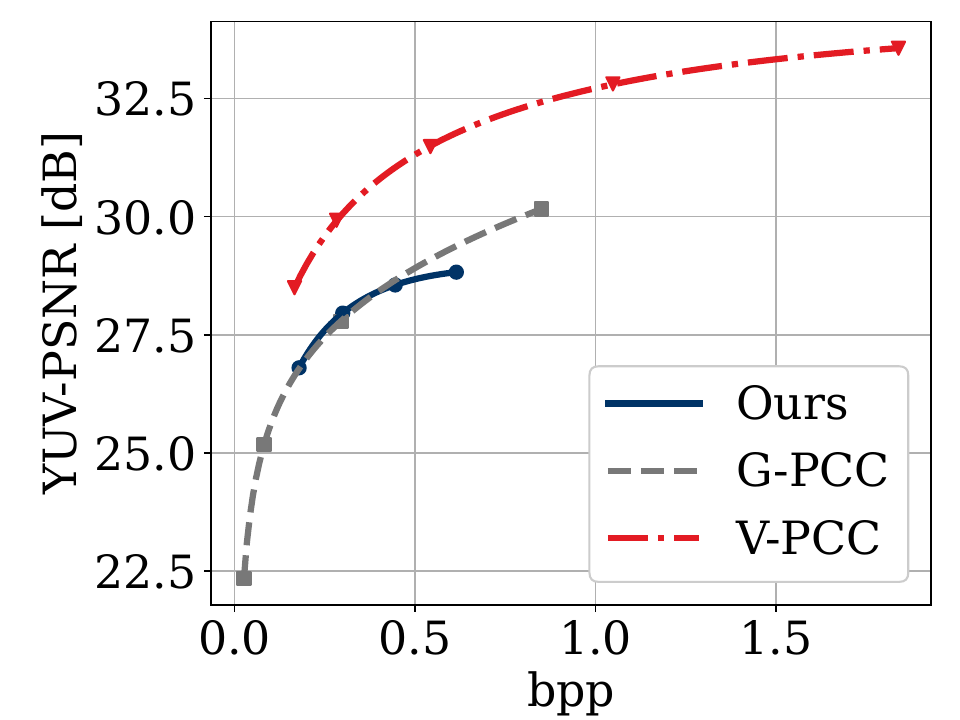}}
    \subfloat{\includegraphics[trim={0cm 0cm 0cm 0cm},clip,width=0.245\textwidth]{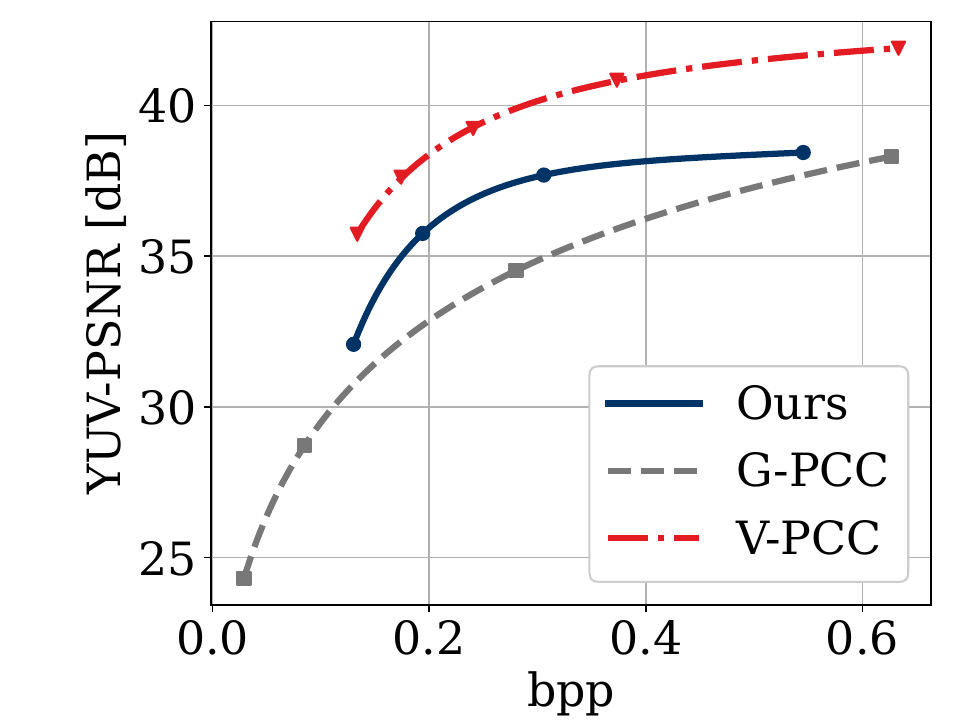}}
    \\
    
    \setcounter{subfigure}{0}
    \subfloat[\textit{phil9}]{\includegraphics[trim={0cm 0cm 0cm 0cm},clip,width=0.245\textwidth]{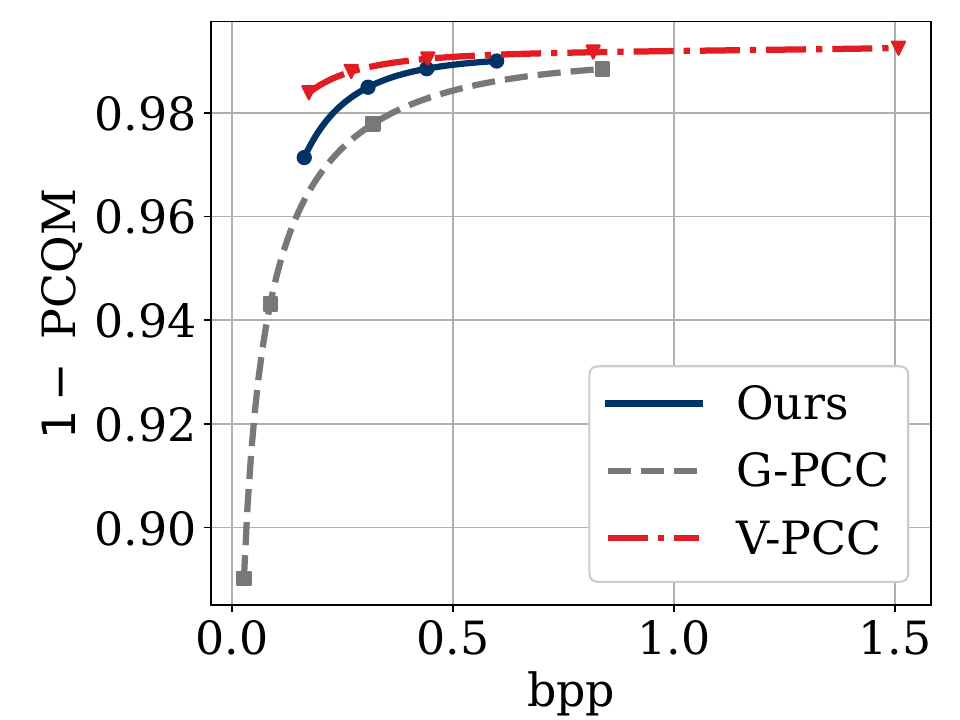} }
    \subfloat[\textit{david9}]{\includegraphics[trim={0cm 0cm 0cm 0cm},clip,width=0.245\textwidth]{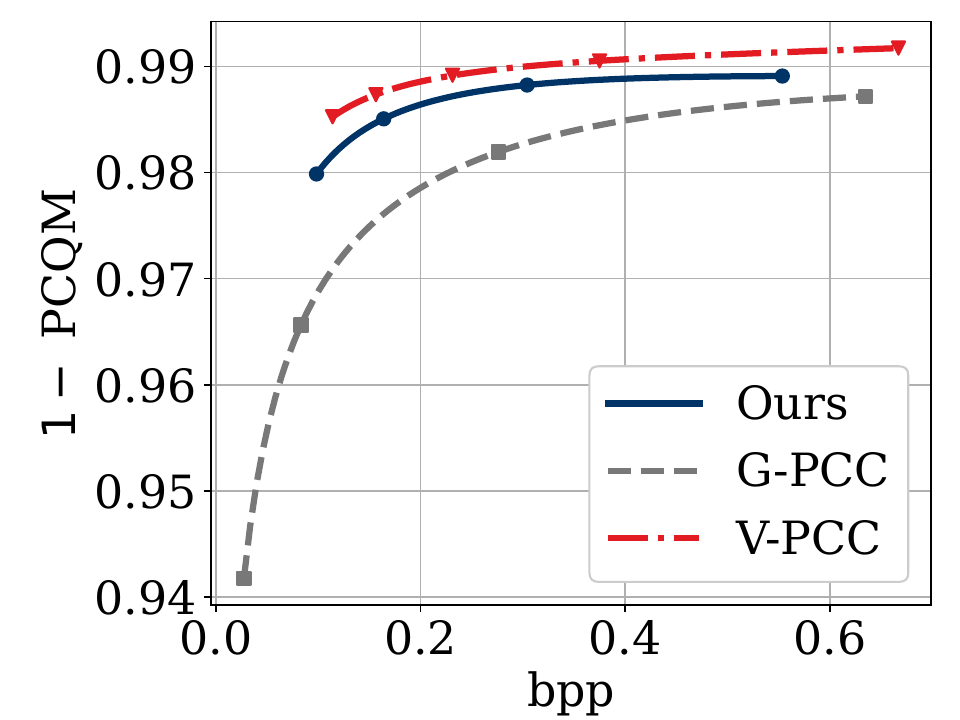}  }
    \subfloat[\textit{andrew9}]{\includegraphics[trim={0cm 0cm 0cm 0cm},clip,width=0.245\textwidth]{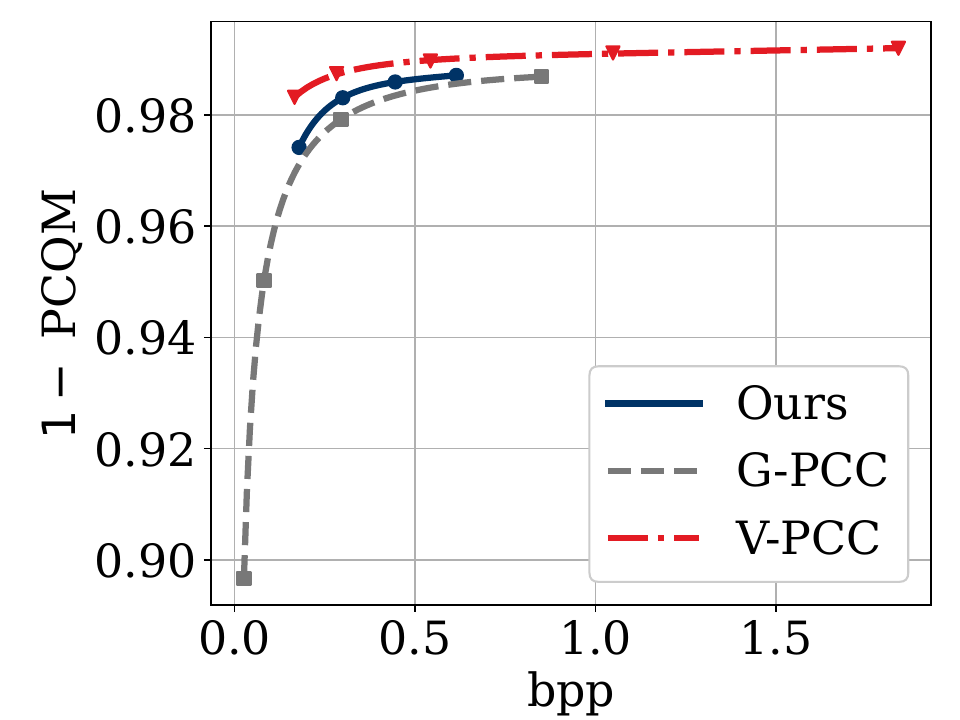} }
    \subfloat[\textit{sarah9}]{\includegraphics[trim={0cm 0cm 0cm 0cm},clip,width=0.245\textwidth]{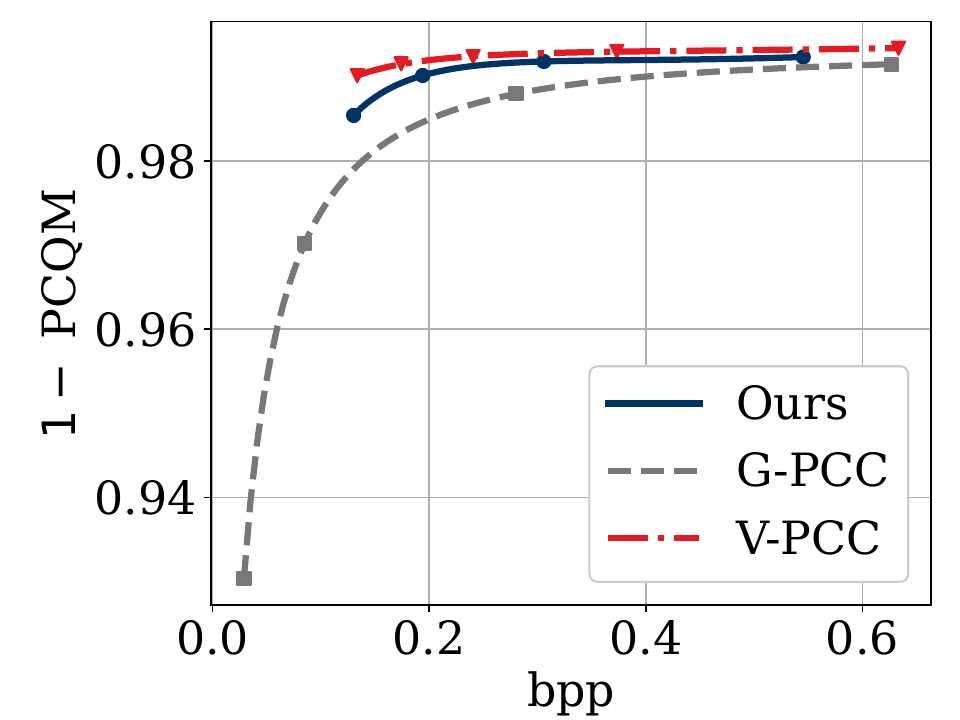} }
    \caption{Rate distortion results on the MVUB~\citep{loop2016microsoft} dataset.}
    \label{fig:appendix_MVUBrd_results}
\end{figure*}

We showcase the rate-distortion results on all point clouds according to the evaluation procedure described in Sec.~\ref{sec:RateDistortion}, as only selected curves were shown in the main document. 
Fig.~\ref{fig:appendix_8ird_results} shows results for the 8iVFBv2 dataset~\citep{upperBodies20178i} and Fig.~\ref{fig:appendix_MVUBrd_results} shows results for the MVUB dataset~\citep{loop2016microsoft}.
Overall, we find increased performance for the geometry metric D1-PSNR over V-PCC~\cite{vpcc2021} and G-PCC~\cite{gpcc2021}. 
For attribute quality our approach achieves a better rate-distortion tradeoff then G-PCC but falls short of the attribute qualities achieved by V-PCC. 
Especially the point cloud \textit{andrew9}, which shows a checkerboard texture pattern on the shirt, is hard to compress using our learned approach. 
Finally, for the joint metric PCQM~\cite{meynet2020pcqm}, we find that our approach consistently achieves higher quality at the given rates compared to G-PCC~\cite{gpcc2021}. 
For higher rate configurations, our method can deliver comparable and sometimes better quality as V-PCC~\cite{vpcc2021}, but falls of when increasing the compression ratio.

Additionally, we notice that the range of compressed bitrates for our model is small compared to the rate-distortion rates achievable by both traditional methods.

\clearpage

\end{document}